\documentclass[10pt,twocolumn,letterpaper]{article}

\usepackage{cvpr}              %

\usepackage[english]{babel}

\usepackage{amsfonts}
\usepackage{amsmath}
\usepackage{amssymb}
\usepackage{amsthm}

\usepackage{xspace}
\usepackage{multirow}
\definecolor{myrowcolor}{rgb}{0.9, 0.95, 1}

\usepackage{comment}
\usepackage[super]{nth}
\usepackage{tikz}
\usepackage{tabularx}
\usepackage{soul}
\usepackage{multirow}
\usepackage{colortbl}
\usepackage{makecell}

\usepackage{layouts}

\usepackage[utf8]{inputenc}
\usepackage{pgfplots}
\DeclareUnicodeCharacter{2212}{−}
\usepgfplotslibrary{groupplots,dateplot}
\usetikzlibrary{patterns,shapes.arrows}
\pgfplotsset{compat=newest}

\usepackage{listings}
\definecolor{delim}{RGB}{20,105,176}
\definecolor{numb}{RGB}{106, 109, 32}
\definecolor{string}{rgb}{0.64,0.08,0.08}
\lstdefinelanguage{json}{
    showspaces=false,
    showtabs=false,
    breaklines=true,
    postbreak=\raisebox{0ex}[0ex][0ex]{\ensuremath{\color{gray}\hookrightarrow\space}},
    breakatwhitespace=true,
    basicstyle=\ttfamily\small,
    upquote=true,
    morestring=[b]",
    stringstyle=\color{string},
    literate=
     *{0}{{{\color{numb}0}}}{1}
      {1}{{{\color{numb}1}}}{1}
      {2}{{{\color{numb}2}}}{1}
      {3}{{{\color{numb}3}}}{1}
      {4}{{{\color{numb}4}}}{1}
      {5}{{{\color{numb}5}}}{1}
      {6}{{{\color{numb}6}}}{1}
      {7}{{{\color{numb}7}}}{1}
      {8}{{{\color{numb}8}}}{1}
      {9}{{{\color{numb}9}}}{1}
      {\{}{{{\color{delim}{\{}}}}{1}
      {\}}{{{\color{delim}{\}}}}}{1}
      {[}{{{\color{delim}{[}}}}{1}
      {]}{{{\color{delim}{]}}}}{1},
}

\newcommand{\X}{\mathcal{X}}
\newcommand{\Y}{\mathcal{Y}}
\newcommand{\T}{\mathcal{T}}
\newcommand{\model}{f}
\newcommand{\task}{T}
\newcommand{\template}{\texttt{temp}}
\newcommand{\captionVal}{c}

\newcommand{\llm}{\Phi_\text{LLM}}

\newcommand{\retrievalVis}{\varphi_\mathtt{vis}}
\newcommand{\retrievalTxt}{\varphi_\mathtt{txt}}
\newcommand{\promptBias}{\pi_\text{bias}}
\newcommand{\promptTemplate}{\pi_\text{template}}
\newcommand{\promptCaption}{\pi_\text{caption}}

\newcommand{\biasSet}{\mathcal{B}}

\newcommand{\biasCategory}{\mathbf{B}}
\newcommand{\biasClass}{b}

\newcommand{\mypar}[1]{\vspace{5pt}\noindent\textbf{#1}}

\newcommand{\dataset}{D}
\newcommand{\database}{\mathcal{D}}

\theoremstyle{definition}

\newcommand{\oursFull}{\textsc{Classifier-to-Bias}\xspace}
\newcommand{\ours}{\textsc{C2B}\xspace}

\definecolor{colorexp}{RGB}{237, 242, 244}
\definecolor{colormethod}{HTML}{c7dcfc}

\newcommand{\inlineColorbox}[2]{\begingroup\setlength{\fboxsep}{1pt}\colorbox{#1}{\hspace*{2pt}\vphantom{Ay}#2\hspace*{2pt}}\endgroup}

\definecolor{ShadedGray}{RGB}{238,238,238}
\definecolor{ModelLightBlue}{RGB}{209, 233, 246}
\definecolor{DrawioBlue}{RGB}{218,232,252}
\definecolor{DrawioOrange}{RGB}{255,230,204}
\definecolor{DrawioGreen1}{RGB}{204,255,204}
\definecolor{DrawioGreen}{RGB}{213,232,212}
\definecolor{DrawioPurple}{RGB}{225,213,231}
\definecolor{DrawioRed}{RGB}{248,206,204}
\definecolor{DrawioYellow}{RGB}{255, 242, 207}
\definecolor{DrawioPink}{RGB}{213, 166, 189}

\newcommand{\rowmethod}{\rowcolor{DrawioOrange}}
\newcommand{\rowb}{\rowcolor{ShadedGray}}

\definecolor{cvprblue}{rgb}{0.21,0.49,0.74}
\usepackage[pagebackref,breaklinks,colorlinks,allcolors=cvprblue]{hyperref}
    
\def\paperID{15887} %
\def\confName{CVPR}
\def\confYear{2025}

\title{\oursFull: Toward Unsupervised Automatic \linebreak Bias Detection for Visual Classifiers}

\author{
    Quentin Guimard$^{1}$ \quad
    Moreno D'Incà$^{1}$ \quad
    Massimiliano Mancini$^{1}$ \quad
    Elisa Ricci$^{1,2}$ \\
    $^1$University of Trento \quad
    $^2$Fondazione Bruno Kessler \\
    {\tt\small \href{https://github.com/mardgui/C2B}{https://github.com/mardgui/C2B}}
}

\begin{document}
\maketitle
\begin{abstract}
A person downloading a pre-trained model from the web should be aware of its biases. 
Existing approaches for bias identification rely on datasets containing labels for the task %
of interest,
something that a non-expert may not have access to, or may not have the necessary resources to collect: this greatly limits the number of tasks where model biases can be identified.
In this work, we present \oursFull (\ours), 
the first bias discovery framework that works \textit{without} access to any labeled data%
: it only relies on a textual description of the classification task to identify biases in the target classification model.
This description is fed to a large language model to generate bias proposals and corresponding captions depicting biases together with task-specific target labels. A %
retrieval model collects images for those captions, which are then used to assess the accuracy of the model w.r.t. the given biases. \ours 
is training-free, does not require any annotations, has no constraints on the list of biases, and can be applied to %
 any pre-trained model on any classification task. Experiments on two publicly available datasets show that \ours discovers biases beyond those of the original datasets and outperforms a recent state-of-the-art bias detection baseline that relies on task-specific annotations, being a promising first step toward addressing task-agnostic unsupervised bias detection.
\end{abstract}
 
\section{Introduction}
\label{sec:intro}

\begin{figure}[!t]
\centering
    \includegraphics[width=1\columnwidth, clip, trim=0.82cm 0.46cm 0.92cm 0.6cm]{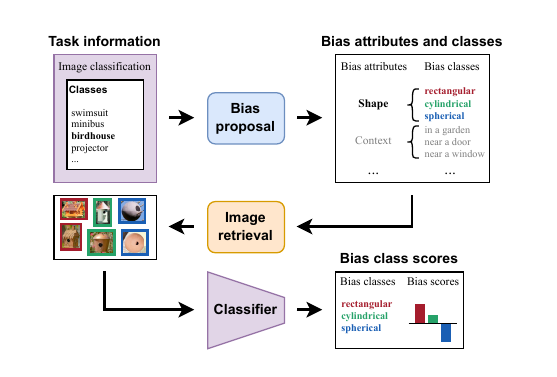}
    \caption{We explore the novel task of bias discovery when we are only given a specific classification task and a pre-trained model. We propose \textbf{\oursFull} (\textbf{\ours}), which automatically detects potential biases in the model, categorizes these biases, and assigns scores to each category.}
    \label{fig:teaser}
\end{figure}

The wide availability of pre-trained machine learning models allows anyone to access them.
A user may search in online libraries (\eg, HuggingFace~\cite{hf}) for models addressing their task of interest, download the most suitable one, and use it for their specific use case. As the user did not develop the model, they may not be aware of potential failure modes~\cite{singla2021understanding,vasudevan2022when,idrissi2023imagenetx} or harmful biases~\cite{bolukbasi2016man,zhao2017men,buolamwini18gender,hendricks2018women, obermeyer2019dissecting,zhao2021understanding}. %
These are usually reported on model cards~\cite{mitchell2019modelcards}, describing how the model has been developed, in which cases it has been tested, its recommended use, and its potential biases. While model cards are a useful tool to describe a model and its shortcomings, (i) a lot of models still come without them; (ii) even if they are present, they may not describe specific cases or biases of interest to the users. This means that a user would need to assess the weaknesses of the downloaded model independently. %

Various tools have already been proposed to measure and mitigate biases in datasets~\cite{bellamy2019ai,wang2022revise} and models~\cite{adebayo2016fairml,bellamy2019ai,cabrera2019fairvis}. While these tools rely on bias labels, recent work has made progress toward open-set failure mode extraction~\cite{deon2022spotlight,eyuboglu2022domino,jain2023distilling,gao2023adaptive,rezaei2024prime} and bias detection~\cite{krishnakumar2021udis,bao2022learning,dinca2024openbias,kim2024b2t,zhang2024discover} without bias labels, even removing the need for a predefined list of biases.
However, in the context of discriminative models such as visual classifiers, all existing methods rely on task-specific labeled data to be able to identify failure modes and measure the biases, limiting the number of tasks where model biases can be identified, and preventing the automation of bias detection.

To address this problem, we present \oursFull (\ours), the first framework that, given a pre-trained visual classifier and its task specification, automatically identifies its biases (\cref{fig:teaser}) in an unsupervised manner, removing the need for task-specific labeled data. We build \ours on two components: a large language model (LLM) and a text-to-image retrieval engine. First, the 
task specifications are fed to the LLM to produce a series of candidate \textit{bias attributes} for the task (\eg, gender, ethnicity), and their corresponding \textit{bias classes} (\eg, male, female). The list of candidate bias classes is fed again to the LLM to generate a set of captions describing both the biases and the task-specific target classes they are associated to. These captions are used by the retrieval module to get images from a large database: the images will constitute our dataset, where, for each image, both the target and bias labels are given by the constructed caption itself. By testing the pre-trained model on this dataset, we can compute the standard bias metrics, verifying which bias class the model is most biased against.%

Importantly, \ours is fully unsupervised and does not suffer from the need of costly annotations, thus it can be applied to automatically detect biases for virtually any classification task.
Moreover, the list of biases is class-specific, as different classification tasks and categories (\eg, birds, deers) may exhibit different biases (\eg, tail color, horns size). Finally, the model is modular and can readily incorporate more powerful LLMs, retrieval approaches, or databases as they become available. We test our framework
on two datasets, \ie, CelebA~\cite{liu2015deep}, and ImageNet-X~\cite{idrissi2023imagenetx}. Results indicate that \ours can uncover biases beyond those annotated in the original datasets, outperforming a recent supervised open-set bias detector~\cite{kim2024b2t}.%

\textbf{Contributions.} To summarize, our contribution is threefold: (i) We propose the new task of
unsupervised bias detection in a pre-trained model given \textit{only} a textual description of the classification task \textit{without} requiring any annotated dataset or manual intervention, and define a protocol to evaluate methods on this setting; (ii) We present \ours, the first framework for this task that uses an LLM to propose a list of biases and generate captions, and multimodal retrieval to produce a dataset from which biases can be estimated; (iii) We show that \ours can discover biases beyond those annotated in the original datasets, being the first step toward addressing this challenging problem and democratizing bias discovery.

\section{Related Work}

\textbf{Bias Detection and Mitigation.}
Bias detection and mitigation are essential to ensure fairness and robustness in machine learning models. Several approaches tackle hidden subgroup imbalances without relying on explicit group annotations \cite{sohoni2020no,liu2021just,li2021discover,singla2021salient}, \eg, by analyzing the mistakes on annotated task data and re-weighting misclassified samples \cite{liu2021just}. %
Other  methods focus on open-set or annotation-free bias discovery \cite{krishnakumar2021udis, li2022discover, zhang2024discover}. UDIS~\cite{krishnakumar2021udis} uses hierarchical clustering %
to surface systematic biases without predefined subgroup definitions. %
DebiAN~\cite{li2022discover} alternates training between a bias discoverer and a classifier to identify and mitigate multiple biases. %
DIM~\cite{zhang2024discover} decomposes feature spaces using Partial Least Squares to uncover multiple unknown biased subgroups and describes them using vision-language models.
More recently, language-driven methods have also gained prominence, using captions and keyword extraction to detect and name potential biases learned by models \cite{kim2024b2t, zhao2024language, ciranni2024say}. B2T~\cite{kim2024b2t} and Zhao et al.~\cite{zhao2024language} 
relies on captions derived from annotated datasets or misclassified samples with known labels, validating these biases using, \eg vision-language similarity. %
SaMyNa~\cite{ciranni2024say} emphasizes interpretability by extracting semantic bias descriptors from large-scale captioning, supporting bias analysis and debiasing. However, all these methods require %
access to annotated, task-specific datasets.

In contrast, our approach, \ours, operates without any kind of annotated data. It relies solely on a natural language description of the classification task and a pre-trained classifier. Using LLM-generated, class-specific bias proposals combined with retrieval-based evaluation, \ours enables fully unsupervised, task-agnostic bias discovery.

Other related methods focus on bias detection or fairness in generative models~\cite{dinca2024openbias,kabra2024gelda,shrestha2024fairrag}.  For instance, OpenBias~\cite{dinca2024openbias} uses LLMs and VQA models to detect and quantify open-set biases in text-to-image generative models, while GELDA~\cite{kabra2024gelda} reveals visual biases in image generators by generating domain-specific attributes with LLMs and annotating them using selected VLMs. %

Inspired by these works, we choose to leverage an LLM as well as a text-to-image retrieval engine to tackle the new task of unsupervised bias detection, where only a pre-trained model and a textual description of the classification task are available, thus eliminating the need for annotated datasets or manual intervention.

\vspace{5pt}
\noindent\textbf{Explainability and Failure Mode Detection.}
A parallel research direction focuses on systematic failure mode discovery and interpretability \cite{agarwal2022estimating,deon2022spotlight,jain2022missingness,hendricks2018generating,augustin2022diffusion}. VoG~\cite{agarwal2022estimating} highlights difficult examples by analyzing gradient variance, and Spotlight~\cite{deon2022spotlight} identifies contiguous underperforming regions in representation space. Saliency-based methods~\cite{jain2022missingness} provide local visual explanations, while counterfactual explanations are generated in either natural language~\cite{hendricks2018generating} or visual form using diffusion models~\cite{augustin2022diffusion}.

Recent work has scaled these efforts using embeddings and generative models. Wiles et al.~\cite{wiles2022discovering} leverages large-scale generative models to automatically find and describe failure clusters. Domino~\cite{eyuboglu2022domino} uses cross-modal embeddings and mixture models to identify and describe coherent data slices where models underperform. Jain et al.~\cite{jain2023distilling} extracts latent directions corresponding to consistent model errors and generates challenging counterfactual examples for data augmentation. AdaVision~\cite{gao2023adaptive} enables interactive discovery of failure modes by combining CLIP-based retrieval with GPT-generated prompts. PRIME~\cite{rezaei2024prime} prioritizes interpretability by starting from human-defined tags and identifying minimal descriptive combinations that explain failures. MAIA~\cite{shaham2024maia} automates interpretability workflows using vision-language models for feature analysis and failure mode discovery. Csurka et al.~\cite{csurka2024could} formalizes language-based error explainability and proposes evaluation metrics to assess the quality of generated explanations.

While these methods analyze model behavior and failure patterns through model outputs or datasets, our work introduces a complementary approach: starting from a high-level task description, \ours uses LLM-based bias proposals and retrieval-based validation to automatically identify, validate, and quantify biases in pre-trained classifiers, enabling large-scale, unsupervised bias discovery across arbitrary classification tasks.

\section{Method}
\label{sec:method}

In this section, we first define bias attributes and bias classes (\cref{sec:def}), and formulate the problem of identifying biases of pre-trained models from their task specifications (\cref{sec:problem}). We then describe \ours, that proposes a list of biases (\cref{sec:method-biases}) and collects data for testing them (\cref{sec:method-retrieval}). Finally, we describe how we measure model's biases from the collected data (\cref{sec:method-measuring}). 

\begin{figure*}[!t]
\centering
    \includegraphics[width=\linewidth, angle=0, origin=c, clip, trim=0.70cm 0.29cm 0.69cm 0.38cm]{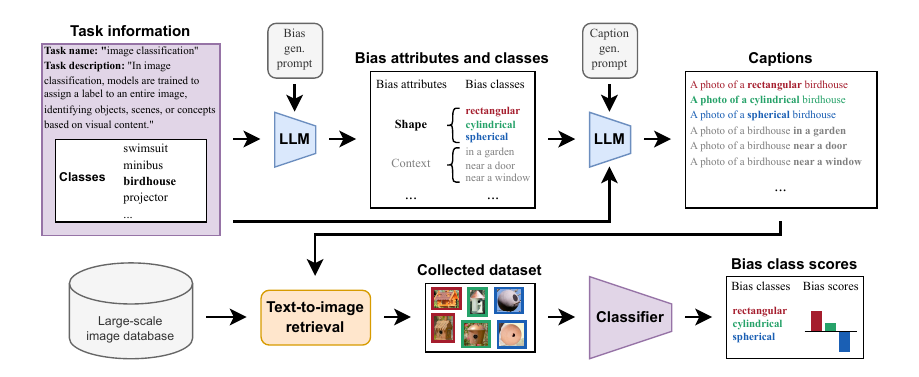}
\caption{\textbf{Overview of \ours}. Given a specific \inlineColorbox{DrawioPurple}{task with an associated description} and a \inlineColorbox{DrawioPurple}{pre-trained classification model}, our approach leverages an \inlineColorbox{DrawioBlue}{LLM} to identify potential candidate bias attributes and corresponding classes. These candidate biases are then used to prompt the LLM to generate captions, which are subsequently utilized by a \inlineColorbox{DrawioOrange}{retrieval} module to collect a dataset of images for bias testing. This dataset enables evaluation of the target classification model, producing scores for each identified bias category. 
}
    \label{fig:method}
\end{figure*}

\subsection{Bias definition}
\label{sec:def}
Formally, we denote the target classification model as $\model:\X\rightarrow\Y$, mapping an image in the space $\X$ to labels in the space $\Y$.
We define as \textbf{bias attributes} $\biasSet=\{\biasCategory_1, \cdots,  \biasCategory_N\}$ the different categories of bias, \ie, factors that can impact its performance~\cite{mehrabi2021survey}.
For each \textbf{bias attribute}, we define a list of \textbf{bias classes} $\biasCategory_i = \{\biasClass_{i,1}, \cdots, \biasClass_{i,n_i}\}$, the different values that each bias attribute can take.
For a given target class, we define the model to be \textbf{biased} or unfair w.r.t specific bias classes depending on the per-class accuracy difference of the model.
Let us denote the per-class accuracy of a model $\model$ on a target class $y$ on data that matches a bias class $\biasClass_{i,j}$ as $A_y(\model, \biasClass_{i,j})$.
We define %
the bias score $\phi_{y,i,j}$ for each bias class $\biasClass_{i,j}$ as
\begin{equation}
    \label{eq:bias-def}
   \phi_{y,i,j} =  A_y(\model, \biasClass_{i,j}) - \frac{1}{n_i-1}\sum_{\biasClass_{i,k}\in \biasCategory_i, k \neq j}{A_y(\model, \biasClass_{i,k}}).
\end{equation}

\subsection{Problem formulation}
\label{sec:problem}

Given a pre-trained model $\model$, our goal is to identify its biases.
We assume $\model$ to be a generic classification model that we did not develop and that we aim to use (\eg, downloaded from the web). Thus, we also assume to have no access to additional data $\model$ had access to (\eg, the dataset set used for training). As we want to avoid costly annotation procedures, we also assume to have no access to any dataset containing labeled images of the classes in $\Y$. 

To address this problem, %
we can rely only on what we know about the model: \ie, the task it has been designed for, the input domain it expects, and the output it predicts. This information can all be expressed using simple, natural language. %
Specifically, we assume to have a general description of the model's task $\task$, the input the model expects to process, and its output (\ie, classes in $\Y$). 

This task is challenging as (i) detecting biases requires measuring the performance of the model on a labeled dataset, potentially containing annotations for the factors of interest and (ii) we have no pre-defined list of biases to start with. This implies that we have to both generate a list of potential biases and to collect a dataset to test them. In the following, we will describe our framework, \ours, that exploits LLMs for proposing biases, and a combination of LLMs and multimodal retrieval for scoring them. An overview of \ours is illustrated in~\cref{fig:method}.

\subsection{Generating bias proposals}
\label{sec:method-biases}
The first step of \ours is to generate a list of bias attributes and classes that we can explicitly test for. To generate the list, we cannot rely on attribute annotations, as we have no access to %
labeled data. Moreover, we cannot use %
lists available online, as %
biases depend on the task of interest.%

To address this problem, we exploit the reasoning capabilities of LLMs. Specifically, let us denote as $\llm$ a generic LLM, mapping textual inputs to textual outputs, \ie, $\llm:\T \rightarrow \T$, with $\T$ being the space of natural language. 
To generate the candidate set of biases, we prompt the LLM to produce a list of bias attributes and classes for the given model and task description.
Because different classes may have different biases,
we query the LLM to give us a list of potential biases $\biasSet^y$ for each class $y\in \Y$. Formally, we have:
\begin{equation}
    \label{eq:bias-gen}
    \biasSet^y = \llm(\promptBias \circ \task \circ y)
\end{equation}
where $\circ$ is the string concatenation operation, $\promptBias$ is the system prompt that instructs the LLM about the task it should perform (\textit{Supp. Mat.}, Appx.~\ref{sec:prompts}), and $\task$ is the model's task description, including input/output specification. 

The output of~\cref{eq:bias-gen}, $\biasSet^y$, is a list\footnote{For simplicity, we assume that the natural language output of the LLM is already formatted as a list.} of bias attributes $\biasSet^y = \{\biasCategory^y_1, \cdots, \biasCategory^y_{N_y} \}$, of size $N_y$. %
We prompt the LLM so that each bias attribute is associated to a list of $n_{y,i}$ bias classes $\biasCategory^y_i = \{\biasClass^y_{i,1}, \cdots, \biasClass^y_{i,n_{y,i}}\}$. %

\subsection{Collecting data}
\label{sec:method-retrieval}
Given the list of candidate bias attributes and classes, we still need a dataset to score them. Ideally, for each target class $y\in\Y$ and bias class $\biasClass^y \in \biasCategory^y$, with $\biasCategory^y\in \biasSet^y$, we would like to have a dataset $\dataset(\biasClass^y, y)$ depicting images of class $y$ exhibiting the bias class $\biasClass^y$, \ie,  $ \dataset(\biasClass^y, y) = \{(x_i,y_i,b^y_i)\}_{i=1}^M$ of size $M$.  Unfortunately, we cannot rely on labeled data for this task as (i) we do not have access to it; (ii) biases change w.r.t. the particular application, thus it is very hard to collect beforehand a dataset containing all possible target and bias classes of interest.  

To address this problem, we resort to external sources. In particular, we assume to have access to a large-scale database $\database$ containing images. Note that $\database$ has no annotations: we just assume it to be large-scale (\ie, potentially web-scale). %
From this database, we can collect images for each target-bias combination of interest using a text-to-image retrieval model. This is achieved in two steps: captions generation and cross-modal retrieval. %

\mypar{Caption generation.} A straightforward solution to create captions is to use a fixed template (\eg, \texttt{an image of [target class] with [bias class]}). However, this would not consider the specificity of the considered task and thus may lead to retrieving erroneous images. For instance, the biases and captions associated to a facial attribute classification task should be different from those in a natural scene classification task.
To avoid this issue, we resort to using the LLM, querying it in two steps: first to create a template for the specific task, and then to %
generate captions specific for target and bias class. 
Formally, we obtain the initial template as:
\begin{equation}
    \label{eq:caption-template}
    \template = \llm(\promptTemplate \circ \task)
\end{equation}
where $\promptTemplate$ is the system prompt that instructs the LLM about the task it should perform (\textit{Supp. Mat.}, Appx.~\ref{sec:prompts}).
Then, we ask the LLM to refine this template for each specific target class $y$ and bias class $b$, as follows:
\begin{equation}
    \label{eq:caption}
    \captionVal^y_{b} = \llm(\promptCaption \circ \template \circ b \circ y).
\end{equation}
where $\captionVal^y_b$ is the resulting caption. %
Note that, while with this strategy we obtain one caption per target-bias class pair, %
in principle we can also generate multiple captions per pair to have more variability, beneficial for more complex tasks.

\mypar{Image retrieval.} The obtained captions can be used to retrieve images from a large scale image database, and we implement the retrieval by using a contrastive vision-language model (VLM) that can score the similarity between a text and image. 
Formally, we can decompose the VLM in three elements: a visual encoder $\retrievalVis:\X\rightarrow \mathbb{R}^d$, a text encoder $\retrievalTxt:\T\rightarrow \mathbb{R}^d$, and a similarity function $\mathtt{sim}:\mathbb{R}^d\times \mathbb{R}^d \rightarrow \mathbb{R}$ (\eg, cosine similarity for CLIP~\cite{radford2021learning}). %
Given a caption, we can %
retrieve a set of images $\mathbf{X}_b^y$ for each class label $y$ and bias $b$  %
as follows:
\begin{equation}
    \label{eq:images-retrieval}
    \mathbf{X}_b^y = \underset{x\in \database}{\text{top-k}}\;\; \mathtt{sim}\left(\retrievalVis(x), \retrievalTxt(\captionVal^y_b) \right) ,
\end{equation}
where we retrieve the $k$ images with the highest matching score.
While these images do not have a ground-truth, they can still help to estimate the biases of the model, with the fidelity of the latter depending on the retrieval accuracy (\cref{sec:eval_setting_3}). %
We also experiment with a different type of %
retrieval system, a Web search engine (details %
in~\cref{sec:exp_prot}).%

\subsection{Measuring biases}
\label{sec:method-measuring}

In the previous section, we have collected a dataset with pseudo-annotations for all bias values and classes of interest. We can now test the target model on this dataset to compute its performance, using the variation across bias classes to %
score each bias.
We use the formula defined in~\cref{eq:bias-def} to score each target-bias pair
relative to other target-bias pairs.
For each bias class, we obtain a bias score between -1 and 1. If the per-class accuracy is higher for a given bias class, the classifier is biased \textit{toward} the bias class (positive bias score). Reciprocally, if the per-class accuracy if lower for a given bias class, the classifier is biased \textit{against} the bias class (negative bias score).
Intuitively, larger absolute bias score values also correspond to stronger biases.

To quantitatively evaluate the quality of the biases detected by \ours, we propose two novel evaluation settings that complement each other. First, in~\cref{sec:eval_setting_1}, we propose to match the detected biases with ground-truth annotations using SBERT embeddings for tasks where datasets with additional bias annotations exist. Second, in~\cref{sec:eval_setting_2}, we propose to assign bias labels (based on the initial proposed biases) to ground-truth task-specific data with a VQA model, to check the agreement between the bias detection method and an expert VQA model.

Collecting scores for all bias attributes and classes enables various types of analyses to explore the biases. Biases can be ranked by score, to assess which attributes are more critical than others. Bias class scores can also be aggregated across bias attributes, to measure which bias attributes exhibit a larger bias. We provide examples of qualitative analyses in~\cref{sec:exp_qual}, and more in the \textit{Supp. Mat.}, Appx.~\ref{sec:additional_qualitative}.%
 
\section{Experiments}
\label{sec:exp}
In this section, we present the results of our experiments, where we apply \ours to detect biases for different models on two different image classification tasks. We first describe our experimental protocol (\cref{sec:exp_prot}), we then quantitatively compare the biases detected by \ours with a state-of-the-art competitor (\cref{sec:exp_sota}), and we provide qualitative examples of detected biases (\cref{sec:exp_qual}). Finally, we analyze the core component of our approach (\cref{sec:eval_setting_3}).

\subsection{Experimental protocol}
\label{sec:exp_prot}

\mypar{Tasks and models.} We test \ours on two different tasks: facial attribute classification and image classification. 

\textit{Facial attribute classification.} We consider the 
CelebA dataset~\cite{liu2015deep}, which is composed of 200k images of celebrity faces annotated with 40 facial attributes, such as \textit{male}, \textit{blond hair}, \textit{young}, or \textit{high cheekbones}. To evaluate \ours, we choose to detect the biases of a recent state-of-the-art model, FaceXFormer~\cite{narayan2024facexformer}. This model relies on a pre-trained Swin/B~\cite{liu2021swin} backbone, coupled with an MLP-based fusion module and an attention-based decoder, which are co-trained on several face datasets. 

\textit{Image classification.} We focus on the widely used %
ImageNet-1K~\cite{ILSVRC15} dataset, %
where each image is associated to one of the 1000 ImageNet classes. We choose to detect the biases of four widely used models, easily accessible through TorchVision~\cite{torchvision2016}, comprising both convolutional networks and transformers: {ResNet50\_V2}~\cite{he2016resnet}, {ResNet101\_V2}~\cite{he2016resnet},  {ResNet152\_V2}~\cite{he2016resnet}, {ViT\_B\_16\_SWAG\_E2E\_V1}~\cite{singh2022swag}.

\mypar{Competitors.}
As we introduce the new problem of unsupervised bias detection relying solely on a textual description of the classification task, there are no existing baseline methods designed for this specific setting.
Nevertheless, we still compare \ours to a recent supervised open-set bias detection method, Bias-to-Text (B2T)~\cite{kim2024b2t}. %
However, unlike \ours which does not requires task-specific labeled data to identify biases in an image classifier, B2T depends on ground-truth task labels to identify model failures. It then employs a pre-trained captioning model to generate descriptions of misclassified images, extracts common keywords, and computes a bias score based on the difference between (i) the similarity (CLIP score) between the discovered bias keyword and misclassified images, and (ii) the similarity between the discovered bias keyword and well-classified images. Thus, we test B2T using the labeled validation sets of the two datasets (\ie, CelebA and ImageNet).%

\mypar{Implementation details.} \ours comprises two components: the LLM that proposes biases and captions, and the retrieval module. For the former, we use Llama 3.1 8B~\cite{dubey2024llama} (see \textit{Supp. Mat.}, Appx.~\ref{sec:llm-comparison} for qualitative LLM comparisons). For the retrieval, we explore two strategies: (i) \textsc{cc12m}: we consider the large Conceptual Captions 12M (CC12M) dataset~\cite{changpinyo2021cc12m} and retrieve images based on CLIP embeddings~\cite{beaumont2022clipretrieval}, and (ii) \textsc{Bing}: instead of using a VLM to retrieve images from an existing database where some image domains may not be well-represented, we propose to retrieve images from the Web using a publicly available search engine, Bing. For both datasets, we retrieve 20 images per caption, and we measure biases as in~\cref{eq:bias-def}, comparing the difference in the model's per-class accuracy between images that match different candidate bias classes. %

\subsection{Quantitative evaluation}
\label{sec:exp_sota}

In this section, we provide a quantitative evaluation to assess the quality of biases identified by \ours. First, we present results based on ground-truth bias annotations, \ie, exploiting attributes annotated in the validation sets of the datasets. %
For biases not in the ground-truth annotations, we introduce an evaluation based on Visual Question Answering (VQA) to measure the relevance of the proposed biases.

\subsubsection{Evaluation with ground-truth annotations}
\label{sec:eval_setting_1}

\mypar{Setting.}
In this setting, we exploit the available attribute annotations, checking if models can detect and quantify those biases correctly. In the case of CelebA, when we focus on a specific attribute classification task (\eg, \textit{blonde hair}) we consider all other 39 attributes (\eg, \textit{young}) as biases measurable with the ground-truth. For ImageNet-1K, we use ImageNet-X~\cite{idrissi2023imagenetx}, \ie, the validation set of ImageNet-1K with 16 labeled ``factors" distinguishing images from typical prototypes (\eg, \textit{orientation}), as %
ground-truth biases. 

To obtain ground-truth biases, we %
test the model on each validation set and check the difference in per-class accuracy among images where the bias attribute is present and images where it is absent. %
We consider a ground-truth bias to be present if this difference is higher than a threshold (\ie, $\tau=0.05$) using the same threshold to assess bias identification on both \ours and B2T scores. We refer to the sign of the bias score as the \textit{direction} of the bias. We provide more results with a varying threshold in the \textit{Supp. Mat.}, Appx.~\ref{sec:additional_quantitative}.

Due to the open-set nature of the detected biases for both \ours and B2T, there are almost no exact matches between them and ground-truth biases. %
To compare the two sets, %
we compute the cosine similarity between SBERT embeddings~\cite{reimers2019sbert} of the ground-truth and detected biases. 
We empirically set the similarity threshold to 0.9 to avoid having too many false positives.
We provide more details and qualitative examples of matches in the \textit{Supp. Mat.}, Appx.~\ref{sec:embedding-matching}.

\mypar{Metrics.} We report results from two complementary perspectives: \mbox{GT~$\rightarrow$~Detected} and \mbox{Detected~$\rightarrow$~GT}. In \mbox{GT~$\rightarrow$~Detected}, we start from each ground-truth bias and check whether it was correctly detected with the same bias direction (\textsc{Hit}), detected with the opposite direction (false hit, \textsc{FH}), or not detected (\textsc{Miss}). This perspective focuses on recall. However, methods that propose many biases may artificially reduce misses. Therefore, we also report \mbox{Detected~$\rightarrow$~GT}, where we start from each detected bias and verify whether it matches a ground-truth bias with the correct direction. This second view measures precision and helps assess the quality of the proposed biases.

\begin{table}[]
    \caption{Proportion (\%) of ground-truth biases detected on CelebA (\mbox{GT~$\rightarrow$~Detected}) and of detected biases corresponding to ground-truth ones (\mbox{Detected~$\rightarrow$~GT}). FH=False Hit.}
    \label{tab:eval_celeba_gt}
    \centering
    \scalebox{0.7}{
    \begin{tabular}{l|ccc |ccc}
    \toprule & \multicolumn{3}{c|}{\textsc{GT} $\rightarrow$ \textsc{Detected}} & \multicolumn{3}{c}{\textsc{Detected} $\rightarrow$ \textsc{GT}}\\
   \textsc{Method} & \textsc{Hit}$(\uparrow)$ & \textsc{FH}$(\downarrow)$ & \textsc{Miss} $(\downarrow)$&\textsc{Hit}$(\uparrow)$ & \textsc{FH} $(\downarrow)$& \textsc{Miss}$(\downarrow)$ \\
    \midrule
      \cellcolor{White} &\multicolumn{6}{c}{\cellcolor{White} \texttt{FaceXFormer}} \\
    \rowb B2T~\cite{kim2024b2t} & 4.53 & \textbf{3.14} & 92.32 & 6.59 & \textbf{4.10} & 89.29\\
     
 \rowmethod \ours (\textsc{Bing}) & \textbf{12.29} & 6.88 & \textbf{80.83}  & \textbf{14.18} & 7.34 & \textbf{78.48} \\
     
  \rowmethod  \ours (\textsc{cc12m})  & 10.76 & 7.75 & 81.49 & 12.75 & 8.15 & 79.09 \\
     
    \bottomrule
    \end{tabular}
    }
\end{table}

\begin{table}[]
    \caption{Proportion (\%) of ground-truth biases detected on ImageNet-X (\mbox{GT~$\rightarrow$~Detected}) and of detected biases corresponding to ground-truth ones (\mbox{Detected~$\rightarrow$~GT}). FH=False Hit.}
    \label{tab:eval_inx_gt}
    \centering
  \scalebox{0.7}{
    \begin{tabular}{l|ccc |ccc}
   \toprule & \multicolumn{3}{c|}{\textsc{GT} $\rightarrow$ \textsc{Detected}} & \multicolumn{3}{c}{\textsc{Detected} $\rightarrow$ \textsc{GT}}\\
   \textsc{Method} & \textsc{Hit}$(\uparrow)$ & \textsc{FH}$(\downarrow)$ & \textsc{Miss} $(\downarrow)$&\textsc{Hit}$(\uparrow)$ & \textsc{FH} $(\downarrow)$& \textsc{Miss}$(\downarrow)$ \\
    \midrule
      \cellcolor{White} &\multicolumn{6}{c}{\cellcolor{White} \texttt{ResNet50\_V2}} \\
     \rowb B2T~\cite{kim2024b2t} & 2.40 & \textbf{2.19} & 95.41 & 0.80 & \textbf{0.71} & 98.49 \\
      
     \rowmethod \ours \textsc{(Bing)} & 7.80 & 8.53 & 83.66 & 2.60 & 2.98 & 94.42 \\
       
     \rowmethod \ours \textsc{(cc12m)} & \textbf{11.18} & 11.30 & \textbf{77.52} & \textbf{2.99} & 2.87 & \textbf{94.14}  \\
   \hline
    \cellcolor{White} &\multicolumn{6}{c}{\cellcolor{White} \texttt{ResNet101\_V2}} \\
    \rowb B2T~\cite{kim2024b2t} & 2.55 & \textbf{2.11} & 95.33 & 0.84 & \textbf{0.69} & 98.47 \\    
    \rowmethod \ours \textsc{(Bing)} & 7.91 & 8.41 & 83.68 & 2.56 & 2.77 & 94.67 \\
      
    \rowmethod \ours \textsc{(cc12m)} & \textbf{11.21} & 11.68 & \textbf{77.11} & \textbf{2.76} & 2.96 & \textbf{94.28}\\
      \hline
       \cellcolor{White} &\multicolumn{6}{c}{\cellcolor{White} \texttt{ResNet152\_V2}} \\
     \rowb B2T~\cite{kim2024b2t} & 2.53 & \textbf{1.72} & 95.75 & 0.83 & \textbf{0.58} & 98.59 \\
      
     \rowmethod \ours \textsc{(Bing)} & 7.53 & 7.93 & 84.54 & 2.75 & 2.79 & 94.46  \\
       
    \rowmethod \ours \textsc{(cc12m)} & \textbf{11.10} & 12.01 & \textbf{76.90} & \textbf{2.81} & 2.95 & \textbf{94.24} \\
   \hline
       \cellcolor{White} &\multicolumn{6}{c}{\cellcolor{White} \texttt{ViT\_B\_16\_SWAG}} \\
   \rowb B2T~\cite{kim2024b2t} & 2.11 & \textbf{1.98} & 95.90  & 0.72 & \textbf{0.67} & 98.61 \\
 
   \rowmethod \ours \textsc{(Bing)} & 7.72 & 7.18 & 85.10 & 2.48 & 2.48 & 95.05\\
      
    \rowmethod  \ours \textsc{(cc12m)} & \textbf{10.85} & 11.07 & \textbf{78.08} & \textbf{2.64} & 2.70 & \textbf{94.66} \\
    \bottomrule
    \end{tabular}
     }
\end{table}

\mypar{Results.}
In~\cref{tab:eval_celeba_gt,tab:eval_inx_gt}, we report the results of our analysis. 
On CelebA (left side of~\cref{tab:eval_celeba_gt}, \mbox{GT~$\rightarrow$~Detected}), \ours outperforms B2T in detecting ground-truth biases, regardless of the retrieval source. We obtain the best results with Bing, detecting between 10.8\% and 12.3\% of the ground-truth biases, compared to only 4.5\% for B2T.
However, B2T exhibits a lower proportion of false hits (3.1\%), whereas \ours shows slightly higher false hit rates (6.9–7.8\%).
This is largely due to retrieval noise and the broader search space introduced by LLM-generated proposals, which can sometimes lead to less precise bias candidates.
Nevertheless, \ours misses fewer ground-truth biases (80.8–81.5\% vs. B2T's 92.3\%).
These trends remain consistent when considering the proportion of detected biases that correspond to ground-truth biases (\mbox{Detected~$\rightarrow$~GT}), where \ours achieves 12.8–14.2\%, compared to 6.6\% for B2T.

On ImageNet-X (\cref{tab:eval_celeba_gt}), across four classifiers, \ours again detects more ground-truth biases than B2T (7.5–11.2\% vs. 2.1–2.6\%), at the cost of higher false hit rates (7.2–12.0\% vs. 1.7–2.2\%). As before, these false hits are attributable to noisy retrieval and the open-ended nature of bias proposals. Still, \ours misses significantly fewer ground-truth biases (76.9–85.1\% vs. 95.3–95.9\%). Interestingly, on ImageNet-X, retrieving from CC12M provides better results than Bing. The \mbox{Detected~$\rightarrow$~GT} evaluation confirms these findings, with a larger fraction of \ours-detected biases aligning with ground-truth ones (2.5–3.0\% vs. 0.6–0.7\% for B2T).

\mypar{Discussion.}
Due to the limited human annotations available in CelebA and ImageNet-X, the proportion of matching biases is low across all methods, particularly on ImageNet-X.
Furthermore, since we are using text embeddings to match the biases, the numbers are also influenced by the semantic proximity between LLM-proposed biases, caption-extracted keywords, and ground-truth annotations. Moreover, a high number of \textsc{Miss} in the \mbox{Detected~$\rightarrow$~GT} case is not necessarily a bad thing, as \ours and B2T may be detecting accurate biases that are not present in the annotation. For these reasons, in the next section we introduce a different evaluation based on visual question answering.%

\subsubsection{VQA-based evaluation}
\label{sec:eval_setting_2}
As ground-truth bias annotations %
may not always be available, 
we also propose an alternative evaluation scheme based on VQA, measuring %
the quality of the proposed biases that are not present in ground-truth annotations.

\mypar{Metric.}
We use a VQA model (LLaVA-1.5-13B~\cite{liu2023llava, liu2023improvedllava}) to %
pseudo-label each image of the validation sets with each of the proposed bias by \ours or B2T. %
Additional details can be found in the \textit{Supp. Mat.}, Appx.~\ref{sec:vqa_details}. From these pseudo-labels, we can use the ground-truth task labels (\ie, class annotations) to check how the per-class accuracy varies over different bias classes, using the same threshold $\tau=0.05$ of \cref{sec:eval_setting_1}. For each bias class $b$ 
proposed by the model, there are three cases: a positive bias is detected ($\phi_b\geq\tau$), a negative bias is detected ($\phi_b\leq-\tau$), or the bias is not detected ($|\phi_b|<\tau$).
For each bias class, we define the agreement score to be 1 if both scores fall in the same case, -1 if they detect opposite biases, and  %
0 otherwise. %

\begin{table}[]
    \caption{Detected biases %
    and VQA agreement on CelebA. }%
    \label{tab:eval2_celeba}
    \centering
    \scalebox{0.7}{
    \begin{tabular}{l|c c c  }
    \toprule
    \textsc{Model}  & \cellcolor{ShadedGray} B2T~\cite{kim2024b2t} &  \cellcolor{DrawioOrange} \ours \textsc{(Bing)} &  \cellcolor{DrawioOrange}  \ours \textsc{(cc12m)} \\
    \midrule
    \texttt{FaceXFormer} 
    & \cellcolor{ShadedGray} 0.10 
   &  \cellcolor{DrawioOrange} 0.22 
  &  \cellcolor{DrawioOrange} \textbf{0.25} \\
    \bottomrule
    \end{tabular}
    }
\end{table}

\begin{table}[]
    \caption{Detected biases %
    and VQA agreement on ImageNet-X. }%
    \label{tab:eval2_inx}
    \centering
    \scalebox{0.7}{
    \begin{tabular}{l|c c c  }
    \toprule
    \textsc{Model}  & \cellcolor{ShadedGray} B2T~\cite{kim2024b2t} &  \cellcolor{DrawioOrange} \ours \textsc{(Bing)} &  \cellcolor{DrawioOrange}  \ours \textsc{(cc12m)} \\
    \midrule
\texttt{ResNet50\_V2}
    & \cellcolor{ShadedGray} 0.20 
   &  \cellcolor{DrawioOrange} \textbf{0.26} 
  &  \cellcolor{DrawioOrange} 0.20 \\
\texttt{ResNet101\_V2}
    & \cellcolor{ShadedGray} 0.19 
   &  \cellcolor{DrawioOrange} \textbf{0.28} 
  &  \cellcolor{DrawioOrange} 0.20 \\
\texttt{ResNet152\_V2}
    & \cellcolor{ShadedGray} 0.20 
   &  \cellcolor{DrawioOrange} \textbf{0.28} 
  &  \cellcolor{DrawioOrange} 0.20 \\
\texttt{ViT\_B\_16\_SWAG}
    & \cellcolor{ShadedGray} 0.20 
   &  \cellcolor{DrawioOrange} \textbf{0.32} 
  &  \cellcolor{DrawioOrange} 0.21 \\
    \bottomrule
    \end{tabular}
    }
\end{table}

\mypar{Results.}
\cref{tab:eval2_celeba,tab:eval2_inx} present the results of our VQA-based evaluation on CelebA and ImageNet-X. On CelebA, \ours achieves higher agreement with VQA pseudo-labels (0.22 for Bing and 0.25 for CC12M) than B2T (0.10), indicating that the biases discovered by \ours are more aligned with visually detectable and semantically coherent bias attributes. On ImageNet-X, while \ours often achieves higher agreement scores than B2T with Bing (\ie, +0.09 on average), results with CC12M are either the same or comparable to those of B2T (\ie, %
+0.005 on average). We attribute this to the challenge of relying on %
retrieval-based data collection, %
without access to annotated failure cases or task-specific captions. In contrast, B2T benefits from captions derived from known misclassifications. Despite this more challenging and fully unsupervised setting, \ours still achieves agreements that are competitive with or superior to B2T. This confirms that \ours's proposals are not only meaningful and visually grounded but also relevant for bias detection in settings where no annotations are available.

\mypar{Discussion.}
The results presented in this section are influenced by both the types of questions posed to the VQA model and the model's own inherent biases. Given these limitations, evaluations based on ground-truth annotations and those derived from VQA outputs serve as complementary approaches. Overall, biases proposed by the LLM tend to offer more relevant bias indicators than keywords extracted from captions as in B2T~\cite{kim2024b2t}, allowing \ours to detect a greater proportion of ground-truth biases and achieve higher agreement with the VQA model than B2T.

\subsection{Qualitative evaluation}
\label{sec:exp_qual}
In this section, we show examples of biases discovered by \ours for different models on different tasks.

\begin{figure*}[ht!]
  \centering
  \includegraphics[width=\textwidth]{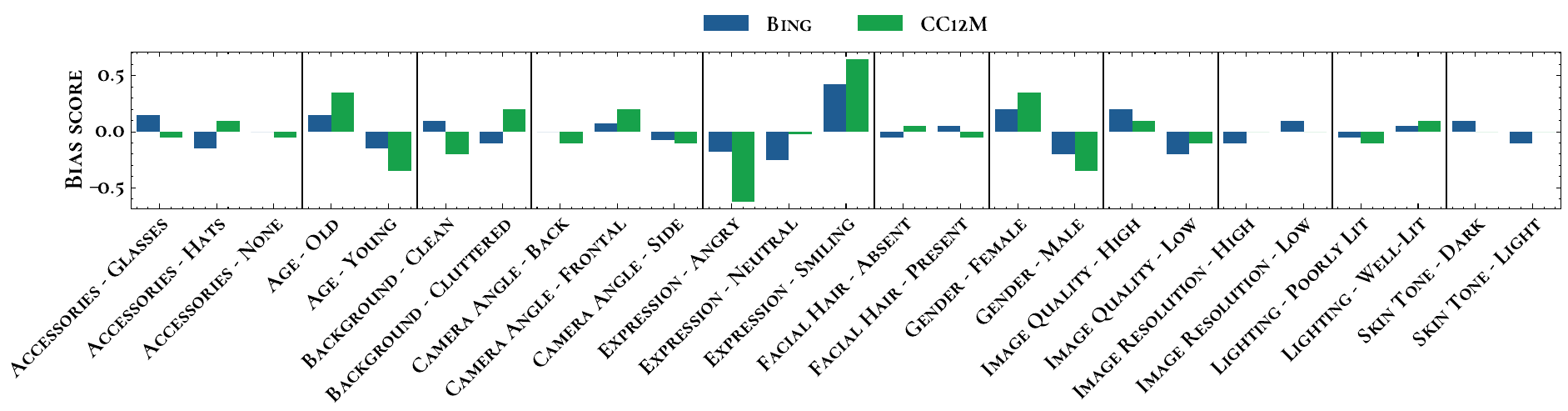}
  \caption{Examples of biases found by \ours for the \textit{high cheekbones} target attribute on CelebA.}
  \label{fig:celeba_qualitative}
\end{figure*}

In~\cref{fig:celeba_qualitative}, we show %
biases that \ours detects for the \textit{high cheekbones} attribute on face attribute classification, %
with bias scores
from different retrieval methods. We can see that a strong bias was detected on the \textit{expression} %
attribute: the classifier appears to perform better when people are smiling and worse when they appear angry. %
Other smaller biases were detected, such as \textit{young}, and \textit{male}.

\begin{figure}[ht!]
  \centering  \includegraphics[width=\columnwidth]{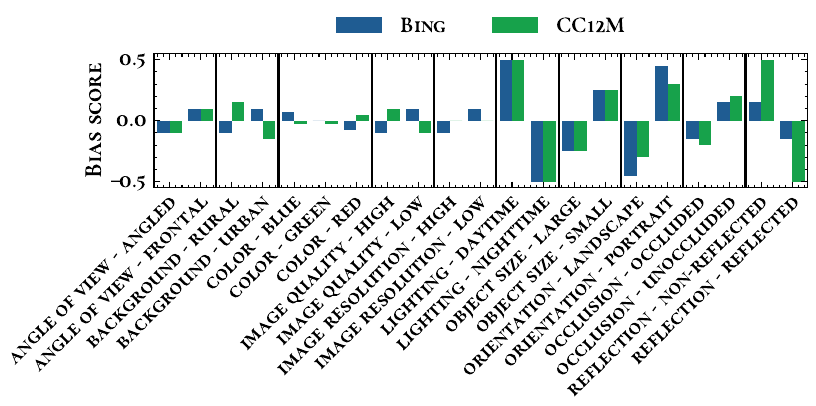}
  \caption{Examples of biases found by \ours for the \textit{minibus} class on ImageNet-X ({ResNet50\_V2}).}
  \label{fig:inx_qualitative_rn50}
\end{figure}

\begin{figure}[t]
  \centering
  \includegraphics[width=\columnwidth]{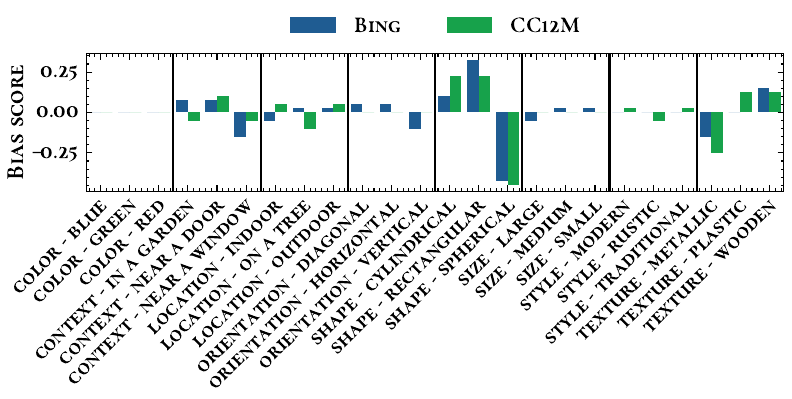}
  \caption{Examples of biases found by \ours for the \textit{birdhouse} class on ImageNet-X ({ViT\_B\_16\_SWAG}).}
  \label{fig:inx_qualitative_vit}
\end{figure}

\cref{fig:inx_qualitative_rn50} shows a similar analysis for biases detected by \ours for the \textit{minibus} class on image classification for the ResNet50\_V2 model. %
Strong biases were detected on the \textit{lighting}, \textit{orientation}, and \textit{presence of reflection} bias attribute: the classifier appears stronger on daytime images, portrait-oriented images, and images without reflections, while %
performing worse on nighttime images, landscape-oriented images, and images with reflections. 
Finally,~\cref{fig:inx_qualitative_vit} depicts biases for the \textit{birdhouse} class again for image classification, but for the ViT\_B\_16\_SWAG model. A strong bias is found on the \textit{shape} bias attribute: the classifier seems to be performing better on images of rectangular birdhouses, while it underperforms on images of spherical birdhouses.

\subsection{Evaluation of VLM-based retrieval}
\label{sec:eval_setting_3}

Retrieving images matching the target classes and biases present in the captions is essential for the detected biases to be trustworthy.
Therefore, %
in this section, we %
analyze the %
CLIP-based retrieval approach used by \ours, by trying to retrieve images for which we have annotations. %
For a VQA-based analysis of both the Bing-based retrieval and CLIP-based retrieval on CC12M, please refer to \textit{Supp. Mat.}, Appx.~\ref{sec:eval_setting_4}.
To perform our analysis, we %
use CLIP to retrieve images from the validation sets of CelebA and ImageNet-X, using handcrafted captions as input: %
combinations of attributes on CelebA, and %
classes and factors on ImageNet-X.

\mypar{Metric.} 
We %
use %
recall@$K$, %
where $K$ varies for each caption to match the number of corresponding ground-truth samples %
in the dataset. %
We show results using recall @ $A \cdot K$, where $A$ is a coefficient that varies between 0.01 and 1.0.

\begin{figure}[t!]
  \centering
  \includegraphics[width=\columnwidth]{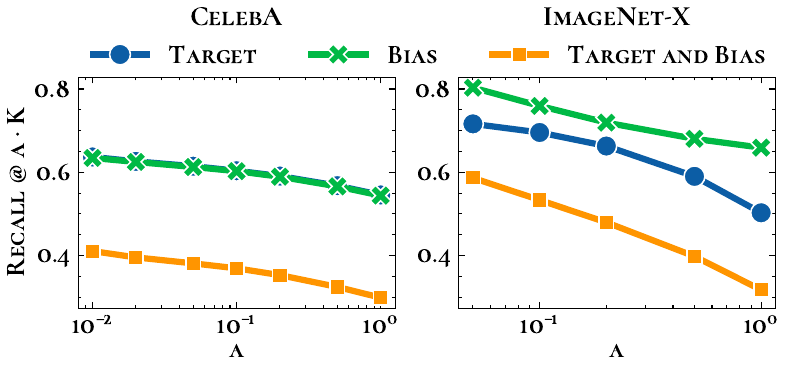}
  \caption{Accuracy of VLM-based retrieval.}
  \label{fig:eval_setting_3}
\end{figure}

\mypar{Results.} We present the results in~\cref{fig:eval_setting_3}. For CelebA (left), %
the recall is the same for both target and bias attributes, as expected, since both are drawn from the same set of attributes. On the same dataset, CLIP struggles to retrieve images with combinations of attributes, with a recall @ $0.01 \cdot K$ just above 40\%. For ImageNet-X (right), CLIP is fairly accurate, especially for low values of $A$, 
with more than 80\% and 70\% recall at $0.05 \cdot K$ to retrieve images with specific biases and classes, respectively. %
Generally, the recall is higher when the number of retrieved images decreases relative to the number of potential matching images present in the dataset, which suggests that, the larger the dataset, the more accurate the retrieval is.

\label{sec:experiments}

\section{Conclusion}

We introduced \oursFull (\ours), a novel method for automatically detecting biases in visual classifiers entirely in an unsupervised manner, without the need for task-specific labeled data. Starting from a simple textual description of the classification task and a pre-trained model, \ours %
generates bias hypotheses, retrieves relevant images to test them, and produces a list of biases with their scores.
Although \ours is the first approach to operate in such challenging %
setting, our results show that it surpasses a recent supervised open-set bias detector %
across two evaluation settings: one based on ground-truth annotations and another on VQA-based pseudo-labels.
We believe our work marks an important step toward automated, open-set, and task-agnostic bias detection.

\section*{Acknowledgments}
This work was sponsored by the MUR PNRR project FAIR - Future AI Research (PE00000013), funded by NextGeneration, and the  %
Italian Ministerial grants PRIN 2022: “B-FAIR: Bias-Free Artificial Intelligence methods for automated visual Recognition” (CUP E53D23008010006), and the EU project ELIAS (No.01120237).

{
    \small
    \bibliographystyle{ieeenat_fullname}
    \bibliography{main}
}

\clearpage
\appendix
\setcounter{page}{1}
\pagestyle{plain}
\onecolumn

\begin{center}
    {\Large \textbf{\oursFull: Toward Unsupervised Automatic\\[0.2em] Bias Detection for Visual Classifiers}}\\[1.2em]
    {\Large Supplementary Material}\\[5.0em]
    {\large \textbf{Table of Contents}}\\[2.0em]
\end{center}

\vspace{1em}
\noindent \cref{sec:limitations}: Limitations of our approach\dotfill~\pageref{sec:limitations}\\[0.5em]
\cref{sec:domain_shift}: Domain shift and retrieval error\dotfill~\pageref{sec:domain_shift}\\[0.5em]
\cref{sec:additional_quantitative}: Additional quantitative results with varying thresholds\dotfill~\pageref{sec:additional_quantitative}\\[0.5em]
\cref{sec:additional_qualitative}: Additional qualitative analyses based on \ours bias scores\dotfill~\pageref{sec:additional_qualitative}\\[0.5em]
\cref{sec:llm_proposed_biases}: Evaluation of the LLM-proposed biases\dotfill~\pageref{sec:llm_proposed_biases}\\[0.5em]
\cref{sec:eval_setting_4}: VQA-based evaluation of the retrieval system\dotfill~\pageref{sec:eval_setting_4}\\[0.5em]
\cref{sec:vqa_details}: Details and comparisons of VQA models\dotfill~\pageref{sec:vqa_details}\\[0.5em]
\cref{sec:retrieval_diversity}: Retrieved images diversity\dotfill~\pageref{sec:retrieval_diversity}\\[0.5em]
\cref{sec:prompts}: Prompts used for LLM bias proposal\dotfill~\pageref{sec:prompts}\\[0.5em]
\cref{sec:llm-comparison}: Comparison between different LLMs for bias proposal\dotfill~\pageref{sec:llm-comparison}\\[0.5em]
\cref{sec:embedding-matching}: Embedding-based bias matching details and examples\dotfill~\pageref{sec:embedding-matching}\\[0.5em]
\cref{sec:gt-biases-visualization}: Ground-truth bias matrices visualizations\dotfill~\pageref{sec:gt-biases-visualization}\\[2em]

\clearpage

\section{Limitations of our approach}
\label{sec:limitations}
As \ours is the first bias detection method in a truly unsupervised setting, it comes with some limitations. We explore these limitations below, focusing on the two core steps of \ours: bias proposal and image retrieval.

\mypar{LLM-based bias proposal.}
Large language models may have limited knowledge, carry their own biases~\cite{omiye2023llm,gallegos2024bias}, and are known to be prone to hallucinations~\cite{ji2023survey,azamfirei2023large,zhang2023siren}. They may propose irrelevant biases or miss biases that could only be found from the data. This limitation stems from our top-down approach, which relies on explicit proposals rather than discovering biases purely from observed failures. In contrast, bottom-up approaches~\cite{krishnakumar2021udis,kim2024b2t,zhao2024language,ciranni2024say} that mine biases from annotated model failures are themselves constrained by the available data and annotations: a model may exhibit biases not represented in the dataset. These two paradigms can be seen as complementary. Additionally, LLM proposals are sensitive to prompting; while we report the prompts used in~\cref{sec:prompts}, alternative prompting strategies or fine-tuned LLMs could yield different or improved results. The modularity of our framework allows for future integration of more specialized or domain-adapted LLMs to improve proposal coverage.

\mypar{Image retrieval.}
The accuracy of \ours strongly depends on the quality of the retrieval system. If retrieved images fail to match the intended target and bias attributes, bias scores will be unreliable. As shown in~\cref{sec:eval_setting_3} and~\cref{sec:eval_setting_4}, current retrieval systems leave room for improvement: across tasks and retrieval sources, less than 50\% of retrieved images correspond to both the intended target and bias classes according to VQA evaluation. This difficulty is due in part to the compositional complexity of the retrieval queries, which vision-language models often struggle with~\cite{thrush2022winoground,hsieh2024sugarcrepe}. Using larger-scale datasets such as DataComp-1B~\cite{gadre2023datacomp} or LAION-5B~\cite{schumann2022laion}, and improved embedding models like SigLIP~\cite{zhai2023sigmoid}, could enhance retrieval precision and compositional understanding. Furthermore, in domain-specific contexts (such as medical imaging), biases may not be directly visible. However, \ours's LLM-based proposal mechanism can surface such biases: for instance, suggesting biases related to hospital type or imaging device. A natural future extension could involve metadata-based retrieval, leveraging contextual attributes beyond visual cues to detect subtle domain-specific biases.

\mypar{Ethical statement and broader impact.}
This work aims to contribute to fairer and more transparent AI by enabling unsupervised detection of biases in pre-trained classifiers. We conduct this research responsibly and with attention to ethical considerations. Nonetheless, due to practical constraints, some socially sensitive attributes (\eg, gender or ethnicity) are treated as closed sets for research purposes only. In addition, \ours inherently reflects the limitations and potential biases of the LLMs and retrieval systems it relies on, and thus may not detect all possible biases. Our intention is not to discriminate against any social group, but rather to raise awareness of the challenges involved in bias discovery and to promote responsible use and auditing of AI models.

\clearpage

\section{Domain shift and retrieval error}
\label{sec:domain_shift}

\ours relies on images retrieved from large-scale external datasets or web sources (\eg, CC12M or Bing). A potential limitation is that these images may differ in style, composition, or distribution compared to the classifier’s training domain (CelebA or ImageNet), introducing domain shift and retrieval noise. This can affect bias scoring accuracy. To better understand the impact of domain shift and retrieval errors, we conducted additional experiments under more controlled conditions.
We considered two alternative retrieval strategies:

\noindent$\bullet$ Retrieval from domain-specific datasets (\textsc{CelebA} and \textsc{ImageNet}): We perform CLIP-based image retrieval directly from the evaluation datasets (CelebA or ImageNet-X) instead of external sources. This reduces domain shift but does not remove retrieval noise, as we still rely on CLIP similarity rather than annotations.

\noindent$\bullet$ Retrieval from labeled subsets (\textsc{CelebA-GT} and \textsc{ImageNet-GT}): We further constrain retrieval to subsets of the datasets where the target class matches the desired target label. This eliminates ambiguity in the target class, though bias attributes are still unknown and must be inferred via CLIP. This setup reduces some noise and simulates a scenario where partial information is available, making it equivalent to the open-set bias detection setting of B2T.

\begin{table}[ht!]
    \caption{Proportion (\%) of ground-truth biases detected on CelebA (GT $\rightarrow$ Detected) and of detected biases corresponding to ground-truth ones (Detected $\rightarrow$ GT). FH=False Hit. Agreement between detected biases and VQA on CelebA.}
    \label{tab:eval_celeba_gt_celeba}
    \centering
    \scalebox{0.7}{
    \begin{tabular}{l|ccc |ccc |c}
    \toprule & \multicolumn{3}{c|}{\textsc{GT} $\rightarrow$ \textsc{Detected}} & \multicolumn{3}{c|}{\textsc{Detected} $\rightarrow$ \textsc{GT}} & \textsc{VQA}\\
   \textsc{Method} & \textsc{Hit}$(\uparrow)$ & \textsc{FH}$(\downarrow)$ & \textsc{Miss} $(\downarrow)$&\textsc{Hit}$(\uparrow)$ & \textsc{FH} $(\downarrow)$& \textsc{Miss}$(\downarrow)$ & \textsc{Agreement} \\
    \midrule
      \cellcolor{White} &\multicolumn{7}{c}{\cellcolor{White} \texttt{FaceXFormer}} \\
    \rowmethod \ours (\textsc{CelebA}) & 11.67 & 7.43 & 80.90 & 12.57 & 10.02 & \textbf{77.41} & 0.27 \\
    \rowmethod \ours (\textsc{CelebA-GT}) & 11.43 & 7.50 & 81.07 & 12.24 & 9.00 & 78.77 & \textbf{0.32} \\
    \bottomrule
    \end{tabular}
    }
\end{table}
\begin{table}[ht!]
    \caption{Proportion (\%) of ground-truth biases detected on ImageNet-X (GT $\rightarrow$ Detected) and of detected biases corresponding to ground-truth ones (Detected $\rightarrow$ GT). FH=False Hit. Agreement between detected biases and VQA on CelebA.}
    \label{tab:eval_inx_gt_in}
    \centering
  \scalebox{0.7}{
    \begin{tabular}{l|ccc |ccc |c}
   \toprule & \multicolumn{3}{c|}{\textsc{GT} $\rightarrow$ \textsc{Detected}} & \multicolumn{3}{c|}{\textsc{Detected} $\rightarrow$ \textsc{GT}} & \textsc{VQA}\\
   \textsc{Method} & \textsc{Hit}$(\uparrow)$ & \textsc{FH}$(\downarrow)$ & \textsc{Miss} $(\downarrow)$&\textsc{Hit}$(\uparrow)$ & \textsc{FH} $(\downarrow)$& \textsc{Miss}$(\downarrow)$ & \textsc{Agreement} \\
    \midrule
       \cellcolor{White} &\multicolumn{7}{c}{\cellcolor{White} \texttt{ResNet50\_V2}} \\
     
     \rowmethod \ours \textsc{(ImageNet)} & 8.25 & 9.61 & 82.13 & 2.73 & 2.86 & 94.41 & 0.28\\
     \rowmethod \ours \textsc{(ImageNet-GT)} & 5.24 & 4.48 & 90.28 & \textbf{3.23} & 2.54 & 94.24 & \textbf{0.40}\\
   \hline
    \cellcolor{White} &\multicolumn{7}{c}{\cellcolor{White} \texttt{ResNet101\_V2}} \\
     
     \rowmethod \ours \textsc{(ImageNet)} & 8.98 & 9.56 & 81.46 & 2.64 & 2.80 & 94.56 & 0.27 \\
     \rowmethod \ours \textsc{(ImageNet-GT)} & 4.09 & 4.28 & 91.63 & \textbf{3.40} & 3.02 & \textbf{93.58} & \textbf{0.44} \\
      \hline
       \cellcolor{White} &\multicolumn{7}{c}{\cellcolor{White} \texttt{ResNet152\_V2}} \\
     
     \rowmethod \ours \textsc{(ImageNet)} & 8.71 & 9.29 & 81.99 & 2.71 & 2.85 & 94.44 & 0.28 \\
     \rowmethod \ours \textsc{(ImageNet-GT)} & 3.39 & 3.49 & 93.11 & \textbf{3.62} & 2.70 & \textbf{93.68} & \textbf{0.47}\\
   \hline
       \cellcolor{White} &\multicolumn{7}{c}{\cellcolor{White} \texttt{ViT\_B\_16\_SWAG}} \\
     
     \rowmethod \ours \textsc{(ImageNet)} & 8.61 & 9.50 & 81.89 & 2.48 & 2.76 & 94.76 & 0.29 \\
     \rowmethod \ours \textsc{(ImageNet-GT)} & 4.07 & 3.81 & 92.12 & \textbf{3.30} & 3.19 & \textbf{93.51} & \textbf{0.48}\\
    \bottomrule
    \end{tabular}
     }
\end{table}

Interestingly, retrieving from the ground-truth dataset itself does not lead to significantly more stable or accurate detection of annotated biases. Bias scores remain comparable to those obtained from external retrieval sources, suggesting that domain shift is not the only source of error. Retrieval noise and the difficulty of capturing subtle bias attributes remain major factors.

Retrieving from labeled subsets results in a trade-off: the hit rate for ground-truth biases decreases, but the false hit rate also reduces, indicating greater precision. Importantly, VQA-based evaluation reveals that both controlled retrieval methods (especially retrieval from labeled subsets) achieve higher agreement with VQA-labeled biases. This suggests that, although \ours detects fewer ground-truth biases in these settings, the detected biases are semantically more meaningful and visually verifiable. These findings highlight that retrieval quality and dataset alignment both impact \ours's performance, and they confirm that part of the observed noise originates from imperfect retrieval. Improving retrieval precision could therefore further enhance bias detection.

\clearpage

\section{Additional quantitative results with varying thresholds}
\label{sec:additional_quantitative}

In~\cref{sec:eval_setting_1}, we chose to present results with a similarity threshold of 0.9 for embedding-based bias matching, and a bias detection threshold of 0.05 for all methods. While we believe that these thresholds represent the best trade-off we could find to detect and match similar biases while avoiding false positives, we present additional results with different threshold values in this section for completeness.

\begin{table}[ht!]
    \parbox{.49\linewidth}{
    \caption{Proportion (\%) of ground-truth biases detected on CelebA (GT $\rightarrow$ Detected) and of detected biases corresponding to ground-truth ones (Detected $\rightarrow$ GT) \textbf{with a similarity threshold of 0.8}. FH=False Hit.}
    \label{tab:eval_celeba_gt_s08}
    \centering
    \scalebox{0.7}{
    \begin{tabular}{l|ccc |ccc}
    \toprule & \multicolumn{3}{c|}{\textsc{GT} $\rightarrow$ \textsc{Detected}} & \multicolumn{3}{c}{\textsc{Detected} $\rightarrow$ \textsc{GT}}\\
   \textsc{Method} & \textsc{Hit}$(\uparrow)$ & \textsc{FH}$(\downarrow)$ & \textsc{Miss} $(\downarrow)$&\textsc{Hit}$(\uparrow)$ & \textsc{FH} $(\downarrow)$& \textsc{Miss}$(\downarrow)$ \\
    \midrule
      \cellcolor{White} &\multicolumn{6}{c}{\cellcolor{White} \texttt{FaceXFormer}} \\
    \rowb B2T~\cite{kim2024b2t} & 8.48 & \textbf{6.52} & 85.00 & 12.25 & \textbf{8.78} & 78.98\\
     
 \rowmethod \ours (\textsc{Bing}) & \textbf{22.69} & 17.26 & \textbf{60.05} & \textbf{27.30} & 18.21 & \textbf{54.49} \\
     
  \rowmethod  \ours (\textsc{cc12m})  & 19.61 & 17.67 & 62.72 & 23.22 & 20.53 & 56.25 \\
     
  \rowmethod \ours (\textsc{CelebA}) & 21.06 & 16.79 & 62.15 & 22.75 & 20.73 & 56.44 \\
    \bottomrule
    \end{tabular}
    }}
    \hfill
    \parbox{.49\linewidth}{

    \caption{Proportion (\%) of ground-truth biases detected on CelebA (GT $\rightarrow$ Detected) and of detected biases corresponding to ground-truth ones (Detected $\rightarrow$ GT) \textbf{with a similarity threshold of 0.95}. FH=False Hit.}
    \label{tab:eval_celeba_gt_s095}
    \centering
    \scalebox{0.7}{
    \begin{tabular}{l|ccc |ccc}
    \toprule & \multicolumn{3}{c|}{\textsc{GT} $\rightarrow$ \textsc{Detected}} & \multicolumn{3}{c}{\textsc{Detected} $\rightarrow$ \textsc{GT}}\\
   \textsc{Method} & \textsc{Hit}$(\uparrow)$ & \textsc{FH}$(\downarrow)$ & \textsc{Miss} $(\downarrow)$&\textsc{Hit}$(\uparrow)$ & \textsc{FH} $(\downarrow)$& \textsc{Miss}$(\downarrow)$ \\
    \midrule
      \cellcolor{White} &\multicolumn{6}{c}{\cellcolor{White} \texttt{FaceXFormer}} \\
    \rowb B2T~\cite{kim2024b2t} & 2.27 & \textbf{0.78} & 96.96 & 3.31 & \textbf{1.14} & 95.55\\
     
 \rowmethod \ours (\textsc{Bing}) & \textbf{6.16} & 2.91 & 90.93 & \textbf{7.82} & 2.70 & 89.48 \\
     
  \rowmethod  \ours (\textsc{cc12m})  & 5.20 & 3.69 & 91.10 & 6.88 & 3.71 & 89.41 \\
     
  \rowmethod \ours (\textsc{CelebA}) & 6.08 & 3.05 & \textbf{90.88} & 6.40 & 5.13 & \textbf{88.46} \\
    \bottomrule
    \end{tabular}
    }}
\end{table}

\begin{table}[ht!]
    \parbox{.49\linewidth}{
  \caption{Proportion (\%) of ground-truth biases detected on ImageNet-X (GT $\rightarrow$ Detected) and of detected biases corresponding to ground-truth ones (Detected $\rightarrow$ GT) \textbf{with a similarity threshold of 0.8}. FH=False Hit.}
  \label{tab:eval_inx_gt_s08}
  \centering
\scalebox{0.7}{
  \begin{tabular}{l|ccc |ccc}
 \toprule & \multicolumn{3}{c|}{\textsc{GT} $\rightarrow$ \textsc{Detected}} & \multicolumn{3}{c}{\textsc{Detected} $\rightarrow$ \textsc{GT}}\\
 \textsc{Method} & \textsc{Hit}$(\uparrow)$ & \textsc{FH}$(\downarrow)$ & \textsc{Miss} $(\downarrow)$&\textsc{Hit}$(\uparrow)$ & \textsc{FH} $(\downarrow)$& \textsc{Miss}$(\downarrow)$ \\
  \midrule
     &\multicolumn{6}{c}{\texttt{ResNet50\_V2}} \\
   \rowb B2T~\cite{kim2024b2t} & 11.34 & \textbf{10.21} & 78.45 & 3.68 & \textbf{3.25} & 93.07 \\
    
   \rowmethod \ours \textsc{(Bing)} & 22.50 & 25.17 & 52.33 & 7.68 & 8.38 & \textbf{83.93} \\
     
   \rowmethod \ours \textsc{(cc12m)} & \textbf{28.06} & 29.28 & \textbf{42.65} & \textbf{7.78} & 7.85 & 84.38  \\
   \rowmethod \ours \textsc{(ImageNet)} & 22.96 & 24.79 & 52.25 & 7.76 & 7.89 & 84.35 \\
 \hline
  &\multicolumn{6}{c}{\texttt{ResNet101\_V2}} \\
  \rowb B2T~\cite{kim2024b2t} & 10.96 & \textbf{9.69} & 79.34 & 3.50 & \textbf{3.18} & 93.32 \\    
  \rowmethod \ours \textsc{(Bing)} & 22.07 & 24.43 & 53.50 & 7.41 & 8.22 & \textbf{84.37} \\
    
  \rowmethod \ours \textsc{(cc12m)} & \textbf{27.59} & 29.62 & \textbf{42.79} & 7.32 & 7.93 & 84.75\\
  \rowmethod \ours \textsc{(ImageNet)} & 23.69 & 24.77 & 51.54 & \textbf{7.50} & 7.98 & 84.53 \\
    \hline
     &\multicolumn{6}{c}{\texttt{ResNet152\_V2}} \\
   \rowb B2T~\cite{kim2024b2t} & 10.91 & \textbf{10.21} & 78.87 & 3.49 & \textbf{3.38} & 93.13 \\
    
   \rowmethod \ours \textsc{(Bing)} & 21.45 & 23.68 & 54.86 & 7.37 & 8.36 & \textbf{84.27}  \\
     
  \rowmethod \ours \textsc{(cc12m)} & \textbf{27.59} & 29.81 & \textbf{42.60} & 7.28 & 7.91 & 84.81 \\
  \rowmethod \ours \textsc{(ImageNet)} & 22.86 & 24.62 & 52.53 & \textbf{7.68} & 7.80 & 84.51 \\
 \hline
     &\multicolumn{6}{c}{\texttt{ViT\_B\_16\_SWAG}} \\
 \rowb B2T~\cite{kim2024b2t} & 10.10 & \textbf{10.19} & 79.71  & 3.23 & \textbf{3.33} & 93.44 \\

 \rowmethod \ours \textsc{(Bing)} & 20.63 & 22.42 & 56.95 & 6.70 & 7.52 & 85.78\\
    
  \rowmethod  \ours \textsc{(cc12m)} & \textbf{27.23} & 29.92 & \textbf{42.85} & 6.90 & 7.43 & \textbf{85.67} \\
  \rowmethod \ours \textsc{(ImageNet)} & 23.21 & 24.80 & 52.00 & \textbf{6.91} & 7.17 & 85.91 \\
  \bottomrule
  \end{tabular}
   }}
    \hfill
    \parbox{.49\linewidth}{

  \caption{Proportion (\%) of ground-truth biases detected on ImageNet-X (GT $\rightarrow$ Detected) and of detected biases corresponding to ground-truth ones (Detected $\rightarrow$ GT) \textbf{with a similarity threshold of 0.95}. FH=False Hit.}
  \label{tab:eval_inx_gt_s095}
  \centering
\scalebox{0.7}{
  \begin{tabular}{l|ccc |ccc}
 \toprule & \multicolumn{3}{c|}{\textsc{GT} $\rightarrow$ \textsc{Detected}} & \multicolumn{3}{c}{\textsc{Detected} $\rightarrow$ \textsc{GT}}\\
 \textsc{Method} & \textsc{Hit}$(\uparrow)$ & \textsc{FH}$(\downarrow)$ & \textsc{Miss} $(\downarrow)$&\textsc{Hit}$(\uparrow)$ & \textsc{FH} $(\downarrow)$& \textsc{Miss}$(\downarrow)$ \\
  \midrule
     &\multicolumn{6}{c}{\texttt{ResNet50\_V2}} \\
   \rowb B2T~\cite{kim2024b2t} & 0.19 & \textbf{0.50} & 99.30 & 0.06 & \textbf{0.15} & 99.79 \\
    
   \rowmethod \ours \textsc{(Bing)} & 2.77 & 2.40 & 94.83 & 0.94 & 0.76 & 98.30 \\
     
   \rowmethod \ours \textsc{(cc12m)} & \textbf{3.96} & 3.49 & \textbf{92.55} & \textbf{1.07} & 0.95 & \textbf{97.98} \\
   \rowmethod \ours \textsc{(ImageNet)} & 2.71 & 2.69 & 94.60 & 0.92 & 0.84 & 98.24 \\
 \hline
  &\multicolumn{6}{c}{\texttt{ResNet101\_V2}} \\
  \rowb B2T~\cite{kim2024b2t} & 0.37 & \textbf{0.54} & 99.09 & 0.12 & \textbf{0.13} & 99.75 \\    
  \rowmethod \ours \textsc{(Bing)} & 2.93 & 2.49 & 94.58 & 0.96 & 0.79 & 98.25 \\
    
  \rowmethod \ours \textsc{(cc12m)} & \textbf{3.96} & 4.00 & \textbf{92.03} & \textbf{0.98} & 1.05 & \textbf{97.97}\\
  \rowmethod \ours \textsc{(ImageNet)} & 3.09 & 2.99 & 93.93 & 0.96 & 0.91 & 98.14 \\
    \hline
     &\multicolumn{6}{c}{\texttt{ResNet152\_V2}} \\
   \rowb B2T~\cite{kim2024b2t} & 0.34 & \textbf{0.33} & 99.33 & 0.11 & \textbf{0.08} & 99.81 \\
    
   \rowmethod \ours \textsc{(Bing)} & 3.06 & 2.45 & 94.49 & \textbf{1.10} & 0.86 & \textbf{98.04}  \\
     
  \rowmethod \ours \textsc{(cc12m)} & \textbf{3.81} & 4.10 & \textbf{92.10} & 0.94 & 1.00 & 98.06 \\
  \rowmethod \ours \textsc{(ImageNet)} & 3.02 & 3.00 & 93.98 & 0.93 & 0.94 & 98.13 \\
 \hline
     &\multicolumn{6}{c}{\texttt{ViT\_B\_16\_SWAG}} \\
 \rowb B2T~\cite{kim2024b2t} & 0.32 & \textbf{0.26} & 99.42 & 0.10 & \textbf{0.08} & 99.82 \\

 \rowmethod \ours \textsc{(Bing)} & 2.84 & 2.35 & 94.81 & 0.91 & 0.78 & 98.31\\
    
  \rowmethod  \ours \textsc{(cc12m)} & \textbf{3.84} & 3.78 & \textbf{92.39} & \textbf{0.94} & 0.90 & \textbf{98.16} \\
  \rowmethod \ours \textsc{(ImageNet)} & 3.06 & 3.06 & 93.88 & 0.80 & 0.89 & 98.30 \\
  \bottomrule
  \end{tabular}
   }}
\end{table}

\begin{table}[ht!]
    \parbox{.49\linewidth}{
    \caption{Proportion (\%) of ground-truth biases detected on CelebA (GT $\rightarrow$ Detected) and of detected biases corresponding to ground-truth ones (Detected $\rightarrow$ GT) \textbf{with a bias detection threshold of 0.01}. FH=False Hit.}
    \label{tab:eval_celeba_gt_d001}
    \centering
    \scalebox{0.7}{
    \begin{tabular}{l|ccc |ccc}
    \toprule & \multicolumn{3}{c|}{\textsc{GT} $\rightarrow$ \textsc{Detected}} & \multicolumn{3}{c}{\textsc{Detected} $\rightarrow$ \textsc{GT}}\\
   \textsc{Method} & \textsc{Hit}$(\uparrow)$ & \textsc{FH}$(\downarrow)$ & \textsc{Miss} $(\downarrow)$&\textsc{Hit}$(\uparrow)$ & \textsc{FH} $(\downarrow)$& \textsc{Miss}$(\downarrow)$ \\
    \midrule
      \cellcolor{White} &\multicolumn{6}{c}{\cellcolor{White} \texttt{FaceXFormer}} \\
    \rowb B2T~\cite{kim2024b2t} & 6.01 & \textbf{3.46} & 90.53 & 10.72 & \textbf{6.07} & 83.20\\
     
 \rowmethod \ours (\textsc{Bing}) & \textbf{13.26} & 9.48 & \textbf{77.26} & \textbf{16.03} & 11.91 & 72.06 \\
     
  \rowmethod  \ours (\textsc{cc12m})  & 12.29 & 9.44 & 78.28 & 15.83 & 12.77 & \textbf{71.40} \\
     
  \rowmethod \ours (\textsc{CelebA}) & 13.07 & 8.76 & 78.17 & 15.50 & 11.84 & 72.65 \\
    \bottomrule
    \end{tabular}
    }}
    \hfill
    \parbox{.49\linewidth}{
    \caption{Proportion (\%) of ground-truth biases detected on CelebA (GT $\rightarrow$ Detected) and of detected biases corresponding to ground-truth ones (Detected $\rightarrow$ GT) \textbf{with a bias detection threshold of 0.1}. FH=False Hit.}
    \label{tab:eval_celeba_gt_d01}
    \centering
    \scalebox{0.7}{
    \begin{tabular}{l|ccc |ccc}
    \toprule & \multicolumn{3}{c|}{\textsc{GT} $\rightarrow$ \textsc{Detected}} & \multicolumn{3}{c}{\textsc{Detected} $\rightarrow$ \textsc{GT}}\\
   \textsc{Method} & \textsc{Hit}$(\uparrow)$ & \textsc{FH}$(\downarrow)$ & \textsc{Miss} $(\downarrow)$&\textsc{Hit}$(\uparrow)$ & \textsc{FH} $(\downarrow)$& \textsc{Miss}$(\downarrow)$ \\
    \midrule
      \cellcolor{White} &\multicolumn{6}{c}{\cellcolor{White} \texttt{FaceXFormer}} \\
    \rowb B2T~\cite{kim2024b2t} & 4.29 & \textbf{0.95} & 94.76 & 5.51 & \textbf{1.37} & 93.12\\
     
 \rowmethod \ours (\textsc{Bing}) & 9.78 & 3.94 & 86.28 & 9.07 & 4.68 & 86.25 \\
     
  \rowmethod  \ours (\textsc{cc12m})  & 9.46 & 6.89 & 83.65 & 11.23 & 4.72 & 84.04 \\
     
  \rowmethod \ours (\textsc{CelebA}) & \textbf{12.73} & 4.27 & \textbf{83.00} & \textbf{11.99} & 6.42 & \textbf{81.58} \\
    \bottomrule
    \end{tabular}
    }}
\end{table}

\begin{table}[ht!]
    \parbox{.49\linewidth}{
  \caption{Proportion (\%) of ground-truth biases detected on ImageNet-X (GT $\rightarrow$ Detected) and of detected biases corresponding to ground-truth ones (Detected $\rightarrow$ GT) \textbf{with a bias detection threshold of 0.01}. FH=False Hit.}
  \label{tab:eval_inx_gt_d001}
  \centering
\scalebox{0.7}{
  \begin{tabular}{l|ccc |ccc}
 \toprule & \multicolumn{3}{c|}{\textsc{GT} $\rightarrow$ \textsc{Detected}} & \multicolumn{3}{c}{\textsc{Detected} $\rightarrow$ \textsc{GT}}\\
 \textsc{Method} & \textsc{Hit}$(\uparrow)$ & \textsc{FH}$(\downarrow)$ & \textsc{Miss} $(\downarrow)$&\textsc{Hit}$(\uparrow)$ & \textsc{FH} $(\downarrow)$& \textsc{Miss}$(\downarrow)$ \\
  \midrule
     &\multicolumn{6}{c}{\texttt{ResNet50\_V2}} \\
   \rowb B2T~\cite{kim2024b2t} & 2.33 & \textbf{2.19} & 95.47 & 0.92 & \textbf{0.80} & 98.28 \\
    
   \rowmethod \ours \textsc{(Bing)} & 11.51 & 12.09 & 76.40 & 3.23 & 3.53 & 93.24 \\
     
   \rowmethod \ours \textsc{(cc12m)} & \textbf{14.32} & 15.02 & \textbf{70.66} & \textbf{3.49} & 3.52 & \textbf{92.99}  \\
   \rowmethod \ours \textsc{(ImageNet)} & 12.05 & 12.06 & 75.89 & 3.44 & 3.26 & 93.30 \\
 \hline
  &\multicolumn{6}{c}{\texttt{ResNet101\_V2}} \\
  \rowb B2T~\cite{kim2024b2t} & 2.58 & \textbf{2.06} & 95.35 & 1.01 & \textbf{0.80} & 98.19 \\    
  \rowmethod \ours \textsc{(Bing)} & 10.67 & 11.80 & 77.53 & 3.10 & 3.29 & 93.61 \\
    
  \rowmethod \ours \textsc{(cc12m)} & \textbf{14.18} & 14.80 & \textbf{71.03} & \textbf{3.27} & 3.53 & \textbf{93.20}\\
  \rowmethod \ours \textsc{(ImageNet)} & 12.02 & 12.31 & 75.67 & 3.20 & 3.31 & 93.49 \\
    \hline
     &\multicolumn{6}{c}{\texttt{ResNet152\_V2}} \\
   \rowb B2T~\cite{kim2024b2t} & 2.53 & \textbf{1.81} & 95.66 & 0.99 & \textbf{0.72} & 98.29 \\
    
   \rowmethod \ours \textsc{(Bing)} & 10.71 & 11.51 & 77.78 & 3.13 & 3.51 & 93.36  \\
     
  \rowmethod \ours \textsc{(cc12m)} & \textbf{14.13} & 15.09 & \textbf{70.78} & \textbf{3.37} & 3.61 & \textbf{93.02} \\
  \rowmethod \ours \textsc{(ImageNet)} & 11.79 & 12.45 & 75.77 & 3.24 & 3.47 & 93.29 \\
 \hline
     &\multicolumn{6}{c}{\texttt{ViT\_B\_16\_SWAG}} \\
 \rowb B2T~\cite{kim2024b2t} & 2.22 & \textbf{2.04} & 95.75 & 0.95 & \textbf{0.86} & 98.19 \\

 \rowmethod \ours \textsc{(Bing)} & 10.10 & 10.38 & 79.52 & 3.07 & 3.09 & 93.84\\
    
  \rowmethod  \ours \textsc{(cc12m)} & \textbf{14.32} & 14.42 & \textbf{71.27} & \textbf{3.36} & 3.40 & \textbf{93.24} \\
  \rowmethod \ours \textsc{(ImageNet)} & 12.02 & 12.31 & 75.68 & 3.20 & 3.30 & 93.50 \\
  \bottomrule
  \end{tabular}
    }}
    \hfill
    \parbox{.49\linewidth}{
  \caption{Proportion (\%) of ground-truth biases detected on ImageNet-X (GT $\rightarrow$ Detected) and of detected biases corresponding to ground-truth ones (Detected $\rightarrow$ GT) \textbf{with a bias detection threshold of 0.1}. FH=False Hit.}
  \label{tab:eval_inx_gt_d01}
  \centering
\scalebox{0.7}{
  \begin{tabular}{l|ccc |ccc}
 \toprule & \multicolumn{3}{c|}{\textsc{GT} $\rightarrow$ \textsc{Detected}} & \multicolumn{3}{c}{\textsc{Detected} $\rightarrow$ \textsc{GT}}\\
 \textsc{Method} & \textsc{Hit}$(\uparrow)$ & \textsc{FH}$(\downarrow)$ & \textsc{Miss} $(\downarrow)$&\textsc{Hit}$(\uparrow)$ & \textsc{FH} $(\downarrow)$& \textsc{Miss}$(\downarrow)$ \\
  \midrule
     &\multicolumn{6}{c}{\texttt{ResNet50\_V2}} \\
   \rowb B2T~\cite{kim2024b2t} & 2.49 & \textbf{2.08} & 95.43 & 0.61 & \textbf{0.51} & 98.88 \\
    
   \rowmethod \ours \textsc{(Bing)} & 3.50 & 4.58 & 91.93 & 1.74 & 2.30 & 95.96 \\
     
   \rowmethod \ours \textsc{(cc12m)} & \textbf{7.06} & 6.52 & \textbf{86.41} & \textbf{2.07} & 2.11 & \textbf{95.81}  \\
   \rowmethod \ours \textsc{(ImageNet)} & 4.69 & 5.34 & 89.97 & 1.93 & 2.03 & 96.03 \\
 \hline
  &\multicolumn{6}{c}{\texttt{ResNet101\_V2}} \\
  \rowb B2T~\cite{kim2024b2t} & 2.93 & \textbf{1.79} & 95.29 & 0.66 & \textbf{0.48} & 98.85 \\    
  \rowmethod \ours \textsc{(Bing)} & 3.88 & 4.26 & 91.86 & 1.89 & 1.96 & 96.15 \\
    
  \rowmethod \ours \textsc{(cc12m)} & \textbf{6.15} & 7.46 & \textbf{86.39} & \textbf{2.13} & 2.32 & \textbf{95.56}\\
  \rowmethod \ours \textsc{(ImageNet)} & 5.13 & 5.11 & 89.76 & \textbf{2.13} & 2.02 & 95.85 \\
    \hline
     &\multicolumn{6}{c}{\texttt{ResNet152\_V2}} \\
   \rowb B2T~\cite{kim2024b2t} & 2.68 & \textbf{1.49} & 95.84 & 0.64 & \textbf{0.38} & 98.98 \\
    
   \rowmethod \ours \textsc{(Bing)} & 3.78 & 4.36 & 91.86 & 1.77 & 2.13 & 96.10  \\
     
  \rowmethod \ours \textsc{(cc12m)} & \textbf{6.09} & 7.18 & \textbf{86.72} & \textbf{1.96} & 2.31 & \textbf{95.73} \\
  \rowmethod \ours \textsc{(ImageNet)} & 4.82 & 5.28 & 89.90 & 1.92 & 1.98 & 96.10 \\
 \hline
     &\multicolumn{6}{c}{\texttt{ViT\_B\_16\_SWAG}} \\
 \rowb B2T~\cite{kim2024b2t} & 2.16 & \textbf{1.95} & 95.89 & 0.56 & \textbf{0.49} & 98.95 \\

 \rowmethod \ours \textsc{(Bing)} & 3.66 & 3.58 & 92.76 & 1.89 & 1.63 & 96.48\\
    
  \rowmethod  \ours \textsc{(cc12m)} & \textbf{6.20} & 6.60 & \textbf{87.20} & \textbf{1.99} & 1.98 & \textbf{96.02} \\
  \rowmethod \ours \textsc{(ImageNet)} & 4.83 & 5.58 & 89.59 & 1.70 & 1.96 & 96.34 \\
  \bottomrule
  \end{tabular}
   }}
\end{table}

First, we present results with a different similarity threshold for embedding-based bias matching. In~\cref{tab:eval_celeba_gt_s08}, we present results on CelebA with a similarity threshold of 0.8. In~\cref{tab:eval_celeba_gt_s095}, we present results on CelebA with a similarity threshold of 0.95. In~\cref{tab:eval_inx_gt_s08}, we present results on ImageNet-X with a similarity threshold of 0.8. In~\cref{tab:eval_inx_gt_s095}, we present results on ImageNet-X with a similarity threshold of 0.95.

Second, we present results with a bias detection threshold. In~\cref{tab:eval_celeba_gt_d001}, we present results on CelebA with a bias detection threshold of 0.01. In~\cref{tab:eval_celeba_gt_d01}, we present results on CelebA with a bias detection threshold of 0.1. In~\cref{tab:eval_inx_gt_d001}, we present results on ImageNet-X with a bias detection threshold of 0.01. In~\cref{tab:eval_inx_gt_d01}, we present results on ImageNet-X with a bias detection threshold of 0.1.

Our results show that \ours's advantage over B2T is consistent across a wide range of threshold values. Lower thresholds increase recall but also lead to more false positives, whereas stricter thresholds reduce false hits at the expense of missing subtle biases. Across all configurations, \ours maintains a higher proportion of ground-truth biases detected, confirming the method’s robustness. In addition, \ours consistently exhibits a lower miss rate than B2T, despite operating in a fully unsupervised setting. These results highlight that our approach is stable and reliable, and that performance is not overly sensitive to hyperparameter choices.

\clearpage

\section{Additional qualitative analyses based on \ours bias scores}
\label{sec:additional_qualitative}

In this section, we propose additional qualitative analyses based on the bias scores assigned by \ours.

\begin{figure}[ht!]
  \centering  \includegraphics[width=0.49\textwidth]{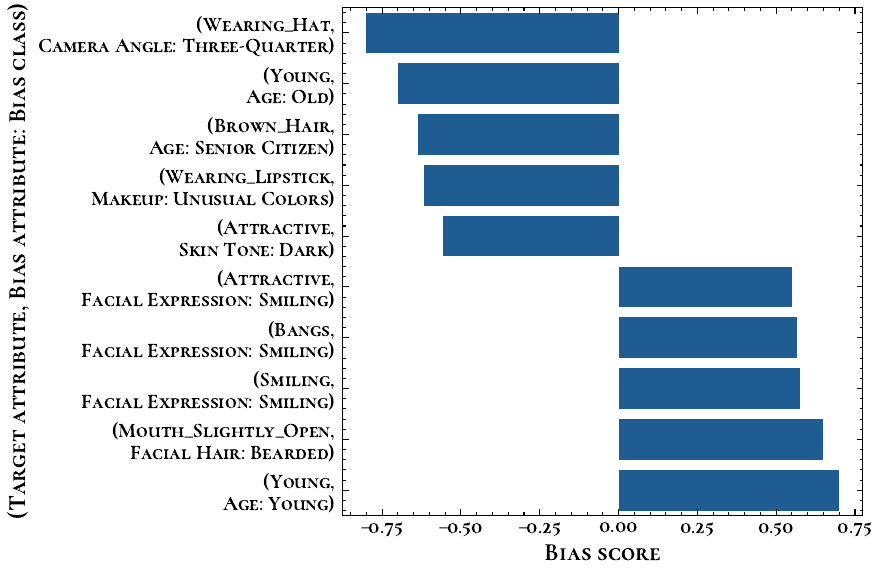}
  \caption{Strongest detected biases of \textbf{FaceXFormer} over all target attributes on \textbf{face attribute classification}.}
  \label{fig:celeba_top_biases}
\end{figure}

\begin{figure}[ht!]
\parbox{.49\linewidth}{
  \centering  \includegraphics[width=0.49\textwidth]{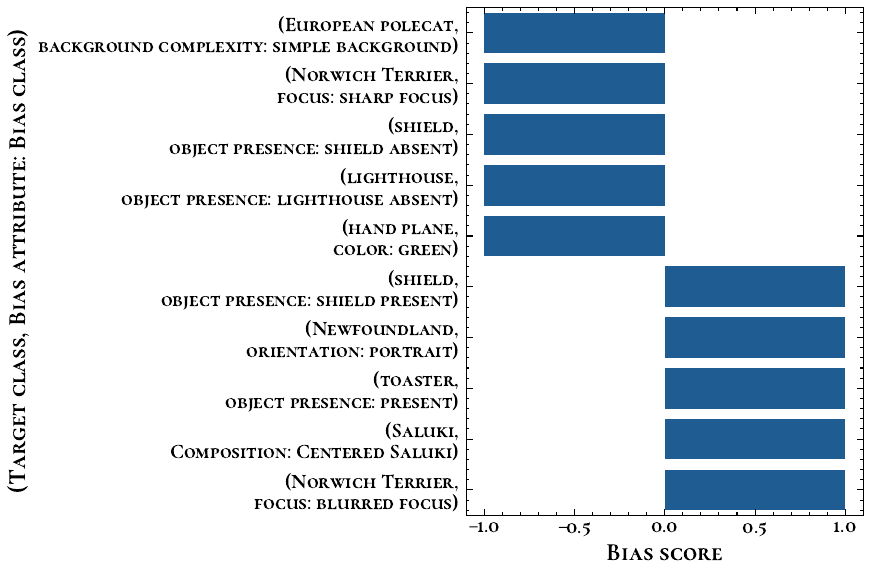}
  \caption{Strongest detected biases of \textbf{ResNet50\_V2} over all target classes on \textbf{image classification.}}
  \label{fig:inx_top_biases_rn50}
  }
\hfill
\parbox{.49\linewidth}{
  \centering  \includegraphics[width=0.49\textwidth]{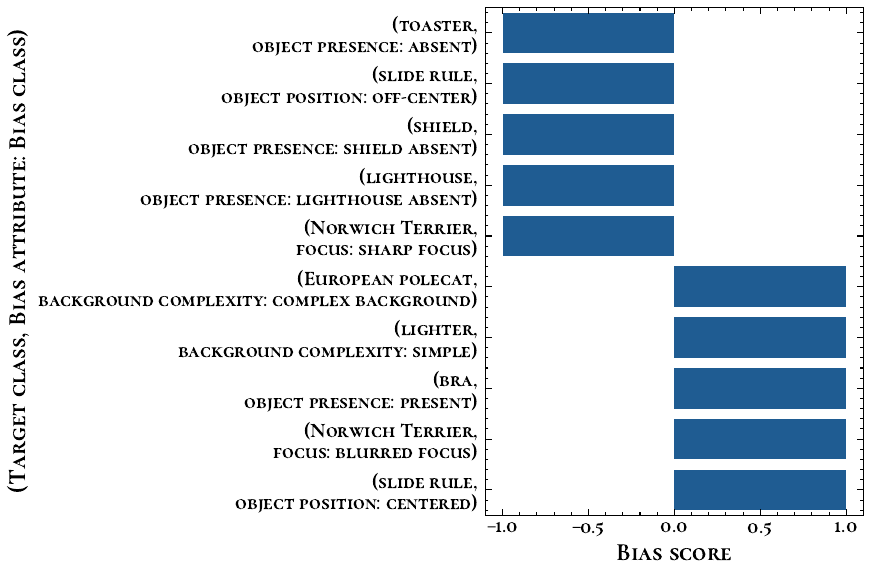}
  \caption{Strongest detected biases of \textbf{ResNet101\_V2} over all target classes on \textbf{image classification.}}
  \label{fig:inx_top_biases_rn101}
  }
\vfill
\parbox{.49\linewidth}{
  \centering  \includegraphics[width=0.49\textwidth]{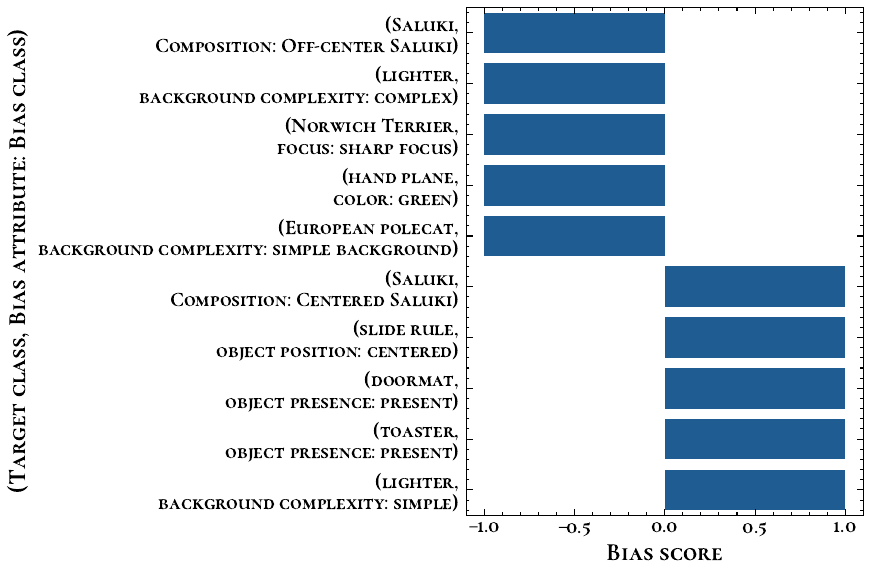}
  \caption{Strongest detected biases of \textbf{ResNet152\_V2} over all target classes on \textbf{image classification.}}
  \label{fig:inx_top_biases_rn152}
  }
\hfill
\parbox{.49\linewidth}{
  \centering  \includegraphics[width=0.49\textwidth]{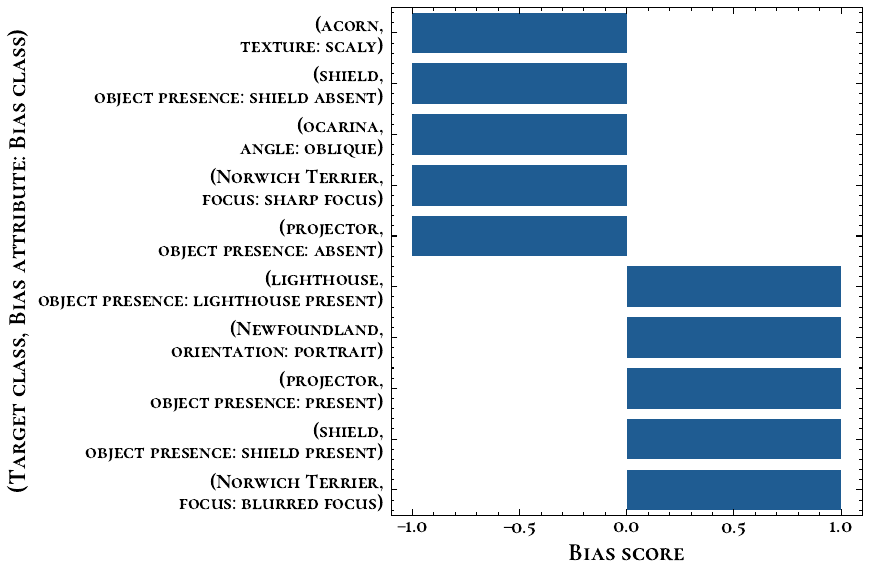}
  \caption{Strongest detected biases of \textbf{ViT\_B\_16\_SWAG} over all target classes on \textbf{image classification.}}
  \label{fig:inx_top_biases_vit}
  }
\end{figure}

\begin{figure}[ht!]
  \centering  \includegraphics[width=0.49\textwidth]{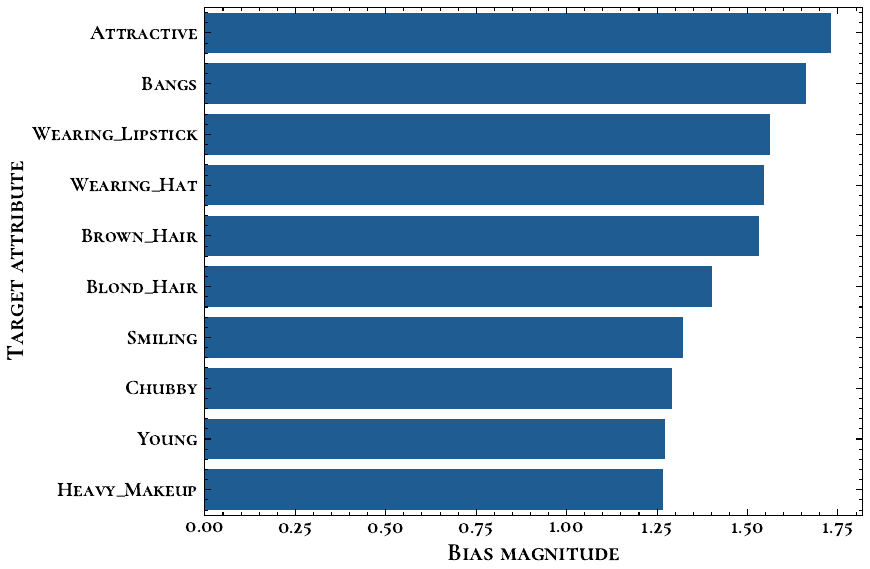}
  \caption{Most bias-affected target attributes for  \textbf{FaceXFormer.}}
  \label{fig:celeba_top_attr}
\end{figure}

\begin{figure}[ht!]
\parbox{.49\linewidth}{
  \centering  \includegraphics[width=0.49\textwidth]{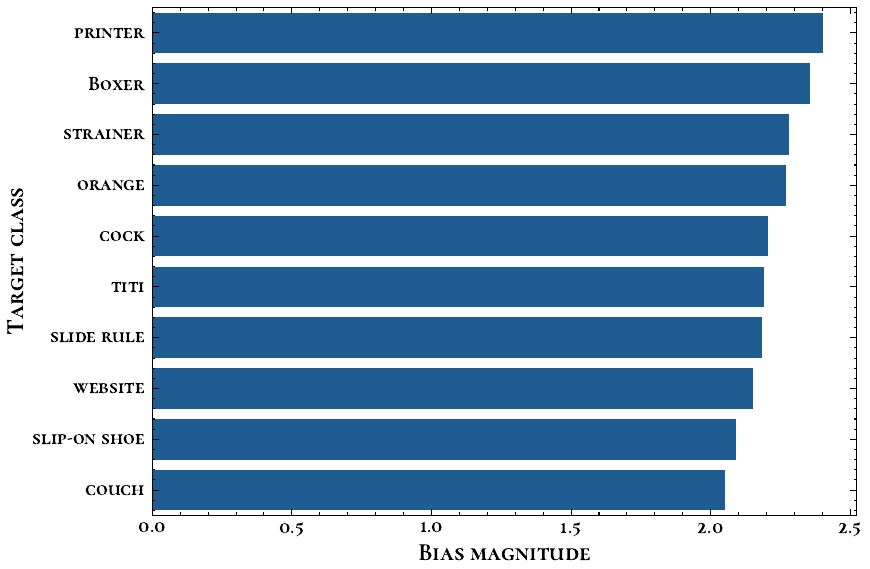}
  \caption{Most bias-affected target classes for \textbf{ResNet50\_V2}.}
  \label{fig:inx_top_classes_rn50}
}
\hfill
\parbox{.49\linewidth}{
  \centering  \includegraphics[width=0.49\textwidth]{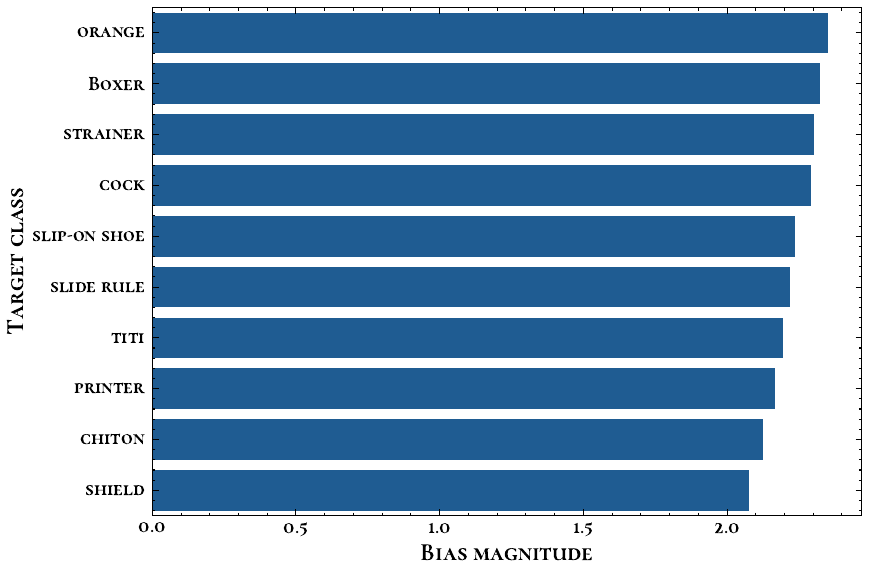}
  \caption{Most bias-affected target classes for \textbf{ResNet101\_V2}.}
  \label{fig:inx_top_classes_rn101}
}
\vfill
\parbox{.49\linewidth}{
  \centering  \includegraphics[width=0.49\textwidth]{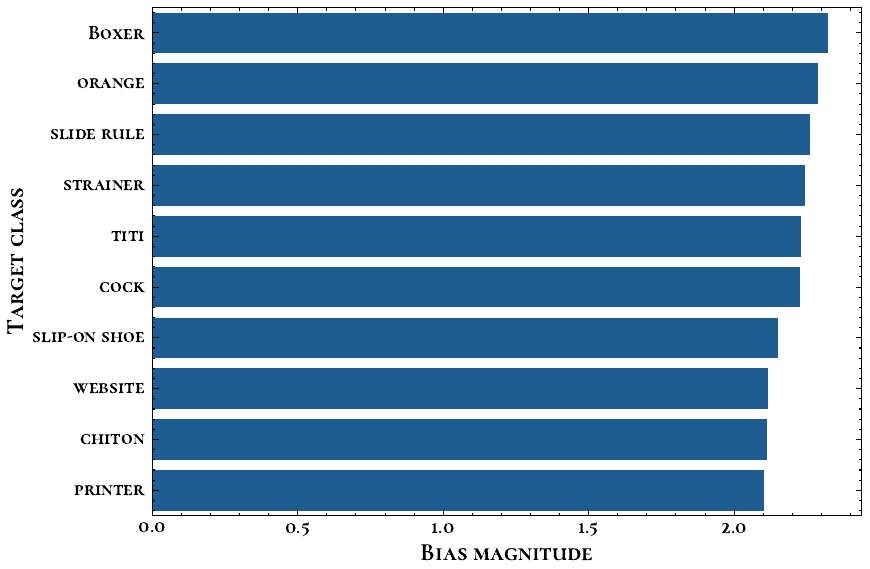}
  \caption{Most bias-affected target classes for \textbf{ResNet152\_V2}.}
  \label{fig:inx_top_classes_rn152}
}
\hfill
\parbox{.49\linewidth}{
  \centering  \includegraphics[width=0.49\textwidth]{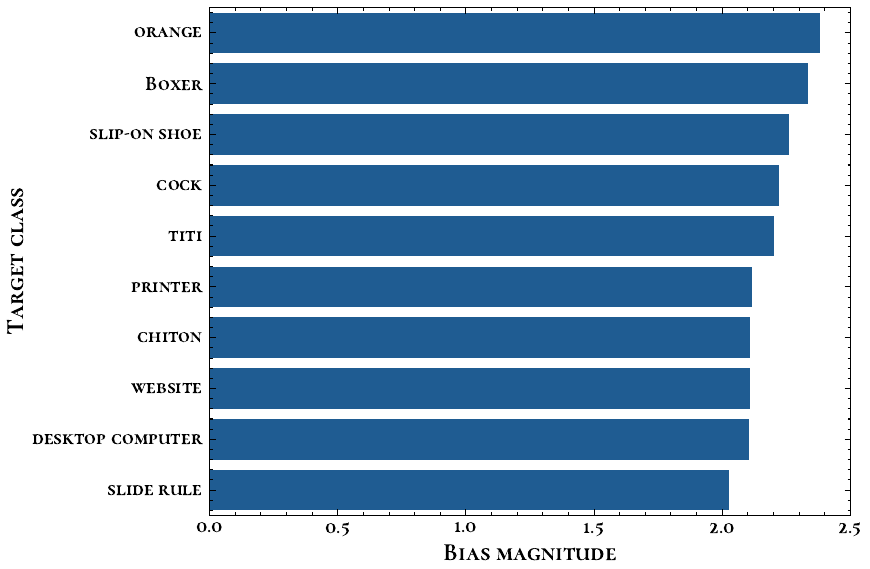}
  \caption{Most bias-affected target classes for \textbf{ViT\_B\_16\_SWAG}.}
  \label{fig:inx_top_classes_vit}
}
\end{figure}

For the first type of analysis, we show the 5 strongest positive biases and the 5 strongest negative biases of a model over all target attributes and classes for a given task.
In~\cref{fig:celeba_top_biases}, we show these biases for FaceXFormer face attribute classification. In ~\cref{fig:inx_top_biases_rn50,fig:inx_top_biases_rn101,fig:inx_top_biases_rn152,fig:inx_top_biases_vit}, we show these biases for {ResNet50\_V2}, {ResNet101\_V2},  {ResNet152\_V2}, and {ViT\_B\_16\_SWAG\_E2E\_V1}, respectively, on image classification.

These figures can be very interesting to inform the user about the strongest biases of a model, and also allow to discover new biases (such as ``camera angle: three-quarter'' for the \textit{wearing hat} attribute in~\cref{fig:celeba_top_biases} or ``color: green'' for the \textit{hand plane} class in~\cref{fig:inx_top_biases_rn50,fig:inx_top_biases_rn152}), but they also illustrate some failure cases of \ours. We can see in~\cref{fig:celeba_top_biases} that ``Age" was proposed as a bias attribute when classifying the \textit{young} attribute, or that ``object presence/absence" was proposed for several different classes in~\cref{fig:inx_top_biases_rn50,fig:inx_top_biases_rn101,fig:inx_top_biases_rn152,fig:inx_top_biases_vit}. These inevitably affect the performance of the classifier, but cannot be considered as ``biases". The implementation of a bias filtering mechanism could lead to improvements in this regard.

For the second type of analysis, we propose to show the target attributes/classes that are the most affected by biases. For each target attribute/class, we define the bias magnitude as the L2 norm of the vector containing the bias scores for all bias classes. In~\cref{fig:celeba_top_attr}, we show the most bias-affected target attributes and their bias magnitude for FaceXFormer face attribute classification. In~\cref{fig:inx_top_classes_rn50,fig:inx_top_classes_rn101,fig:inx_top_classes_rn152,fig:inx_top_classes_vit}, we show the most bias-affected target classes and their bias magnitude for {ResNet50\_V2}, {ResNet101\_V2},  {ResNet152\_V2}, and {ViT\_B\_16\_SWAG\_E2E\_V1}, respectively, on image classification.

These figures can be useful to inform the user about which target attributes/classes are the most affected by biases when using a specific model. For instance, we can see in~\cref{fig:celeba_top_biases} that the \textit{attractive} target attribute is the most affected by biases. From this, the user may want to look at the various bias scores for the different bias attributes and classes that were proposed for the \textit{attractive} target attribute. We can also see in~\cref{fig:inx_top_classes_rn50,fig:inx_top_classes_rn101,fig:inx_top_classes_rn152,fig:inx_top_classes_vit} that \textit{boxer} and \textit{orange} are two of the most bias-affected classes across all four tested models.

\clearpage

\section{Evaluating LLM-proposed biases}
\label{sec:llm_proposed_biases}

\ours relies on large language models (LLMs) to propose candidate bias attributes based on a textual description of the task and target classes. While this approach enables an unsupervised and task-agnostic bias discovery process, it also raises the question of how well the LLM-proposed biases align with known ground-truth biases and whether they contribute to discovering novel biases.
We analyze two aspects:

\noindent$\bullet$ Coverage of ground-truth biases (\textsc{LLM Miss}): We measure the proportion of ground-truth biases that are included among the LLM-proposed biases.

\noindent$\bullet$ Novel bias discovery (\textsc{New Bias}): We evaluate what proportion of detected biases (\ie, biases with high bias scores identified by \ours) are not part of the annotated ground-truth set, indicating the discovery of previously unannotated biases.

\begin{table}[ht!]
    \caption{Proportion (\%) of ground-truth biases detected on CelebA (GT $\rightarrow$ Detected) and of detected biases corresponding to ground-truth ones (Detected $\rightarrow$ GT). FH=False Hit.}
    \label{tab:eval_celeba_gt_llm}
    \centering
    \scalebox{0.7}{
    \begin{tabular}{l|cccc |cccc}
    \toprule & \multicolumn{4}{c|}{\textsc{GT} $\rightarrow$ \textsc{Detected}} & \multicolumn{4}{c}{\textsc{Detected} $\rightarrow$ \textsc{GT}}\\
   \textsc{Method} & \textsc{Hit}$(\uparrow)$ & \textsc{FH}$(\downarrow)$ & \textsc{Retrieval Miss}$(\downarrow)$& \textsc{LLM Miss}$(\downarrow)$&\textsc{Hit}$(\uparrow)$ & \textsc{FH} $(\downarrow)$& \textsc{Not a Bias}$(\downarrow)$& \textsc{New Bias}$(\uparrow)$ \\
    \midrule
      \cellcolor{White} &\multicolumn{8}{c}{\cellcolor{White} \texttt{FaceXFormer}} \\
\rowmethod \ours (\textsc{Bing}) & \textbf{12.29} & 6.88 & \textbf{9.94}  & 70.89 & \textbf{14.18} & 7.34 & \textbf{9.58} & \textbf{68.91} \\
     
\rowmethod \ours (\textsc{cc12m})  & 10.76 & 7.75 & 10.60 & 70.89  & 12.75 & 8.15 & 12.08 & 67.02 \\
     
    \rowmethod \ours (\textsc{CelebA}) & 11.67 & 7.43 & 10.01 & 70.89  & 12.57 & 10.02 & 9.97 & 67.44 \\
    \bottomrule
    \end{tabular}
    }
\end{table}

\begin{table}[ht!]
    \caption{Proportion (\%) of ground-truth biases detected on ImageNet-X (GT $\rightarrow$ Detected) and of detected biases corresponding to ground-truth ones (Detected $\rightarrow$ GT). FH=False Hit.}
    \label{tab:eval_inx_gt_llm}
    \centering
  \scalebox{0.7}{
    \begin{tabular}{l|cccc |cccc}
   \toprule & \multicolumn{4}{c|}{\textsc{GT} $\rightarrow$ \textsc{Detected}} & \multicolumn{4}{c}{\textsc{Detected} $\rightarrow$ \textsc{GT}}\\
   \textsc{Method} & \textsc{Hit}$(\uparrow)$ & \textsc{FH}$(\downarrow)$ & \textsc{Retrieval Miss}$(\downarrow)$& \textsc{LLM Miss}$(\downarrow)$&\textsc{Hit}$(\uparrow)$ & \textsc{FH} $(\downarrow)$& \textsc{Not a Bias}$(\downarrow)$& \textsc{New Bias}$(\uparrow)$ \\
    \midrule
      \cellcolor{White} &\multicolumn{8}{c}{\cellcolor{White} \texttt{ResNet50\_V2}} \\
      
     \rowmethod \ours \textsc{(Bing)} & 7.80 & 8.53 & 16.79 & 66.88 & 2.60 & 2.98 & \textbf{11.81} & \textbf{82.61}\\
       
     \rowmethod \ours \textsc{(cc12m)} & \textbf{11.18} & 11.30 & \textbf{10.65} & 66.88 & \textbf{2.99} & 2.87 & 12.54 & 81.60  \\
       
     \rowmethod \ours \textsc{(ImageNet)} & 8.25 & 9.61 & 15.25 & 66.88 & 2.73 & 2.86 & 12.38 & 82.03  \\
   \hline
   \cellcolor{White} &\multicolumn{8}{c}{\cellcolor{White} \texttt{ResNet101\_V2}} \\

    \rowmethod \ours \textsc{(Bing)} & 7.91 & 8.41 & 17.01 & 66.67 & 2.56 & 2.77 & \textbf{12.10} & \textbf{82.58} \\
      
    \rowmethod \ours \textsc{(cc12m)} & \textbf{11.21} & 11.68 & \textbf{10.45} & 66.67 & \textbf{2.76} & 2.96 & 12.87 & 81.41\\
      
    \rowmethod \ours \textsc{(ImageNet)} & 8.98 & 9.56 & 14.80 & 66.67 & 2.64 & 2.80 & 12.37 & 82.19\\
      \hline
      \cellcolor{White} &\multicolumn{8}{c}{\cellcolor{White} \texttt{ResNet152\_V2}} \\
      
     \rowmethod \ours \textsc{(Bing)} & 7.53 & 7.93 & 17.77 & 66.77 & 2.75 & 2.79 & \textbf{11.56} & \textbf{82.90}  \\
       
    \rowmethod \ours \textsc{(cc12m)} & \textbf{11.10} & 12.01 & \textbf{10.12} & 66.77 & \textbf{2.81} & 2.95 & 12.83 & 81.42 \\
      
    \rowmethod \ours \textsc{(ImageNet)} & 8.71 & 9.29 & 15.22 & 66.77 & 2.71 & 2.85 & 12.33 & 82.11\\
   \hline
     \cellcolor{White}  &\multicolumn{8}{c}{\cellcolor{White} \texttt{ViT\_B\_16\_SWAG}} \\

   \rowmethod \ours \textsc{(Bing)} & 7.72 & 7.18 & 18.03 & 67.07 & 2.48 & 2.48 & \textbf{11.71} & \textbf{83.34}\\
      
    \rowmethod \ours \textsc{(cc12m)} & \textbf{10.85} & 11.07 & \textbf{11.02} & 67.07 & \textbf{2.64} & 2.70 & 13.20 & 81.46 \\
      
    \rowmethod \ours \textsc{(ImageNet)} & 8.61 & 9.50 & 14.83 & 67.07 & 2.48 & 2.76 & 12.62 & 82.15 \\
    \bottomrule
    \end{tabular}
     }
\end{table}

As shown in~\cref{tab:eval_celeba_gt_llm,tab:eval_inx_gt_llm}, the LLM proposals cover approximately 29\% of the ground-truth biases on CelebA and 33\% on ImageNet-X. While this partial coverage reflects the inherent limitations of relying on a language model without task-specific supervision, it also confirms that LLMs can generate relevant and meaningful bias candidates in a wide variety of domains.

At the same time, a large fraction of the biases detected by \ours are not present in the ground-truth annotations: around 68\% on CelebA and 82\% on ImageNet-X. This demonstrates \ours’s ability to discover novel, previously unannotated biases. Some of these newly surfaced biases are confirmed through qualitative inspection and VQA-based validation to reflect real, systematic model behaviors. These results highlight that while LLM proposals do not exhaustively cover all known biases, they provide a rich and diverse starting point for unsupervised bias discovery, enabling \ours to go beyond closed-set annotations and uncover previously overlooked spurious correlations.

\clearpage

\section{VQA-based evaluation of the retrieval system}
\label{sec:eval_setting_4}

As mentioned in~\cref{sec:limitations}, \ours critically depends on the accuracy of the retrieval. In~\cref{sec:eval_setting_3}, we propose to measure the accuracy of the CLIP-based retrieval by retrieving on a labeled dataset and measuring the recall @ K. While this is the only type of possible evaluation using ground-truth annotations, this is not perfectly representative of our use case, because the captions that were used could not contain the LLM-proposed biases, but had to rely on labeled attributes. Moreover, this evaluation can only measure the accuracy of the CLIP-based retrieval, as Bing cannot be used to retrieve images from a local database.

For these reasons, we propose to evaluate the actual retrieved data with a visual question answering (VQA) model. This allows to measure the quality of the retrieved images, by asking the VQA if the attributes that we want are indeed in the images (both the target and bias classes). This also allows a direct comparison between Bing and the CLIP-based retrieval (both on CC12M and the evaluation dataset itself, as seen in~\cref{sec:domain_shift}). We present our results in~\cref{tab:eval4_fac} for the face attribute classification task, and in~\cref{tab:eval4_ic} for the image classification task.

\begin{table}[ht!]
    \parbox{.49\linewidth}{
    \caption{Accuracy of the retrieval according to the VQA on the face attribute classification task.}
    \label{tab:eval4_fac}
    \centering
      \scalebox{0.7}{
  \begin{tabular}{l|ccc}
    \toprule
   \multirow{2}{*}{\textsc{Retrieval method}}  & \multicolumn{3}{c}{\textsc{Accuracy}} \\
   &\textsc{Target}&\textsc{Bias}&\textsc{Both}\\
    \midrule
   \textsc{Bing} & \textbf{78.33} & 60.33 & {46.49} \\
     \textsc{CLIP + CC12M}& 67.72 & {63.69}& 42.43\\
     \textsc{CLIP + CelebA} & 72.98 & \textbf{65.37} & \textbf{46.65}\\
    \bottomrule
    \end{tabular}}}
    \hfill
    \parbox{.49\linewidth}{
    \caption{Accuracy of the retrieval according to the VQA on the image classification task.}
    \label{tab:eval4_ic}
    \centering
      \scalebox{0.7}{
    \begin{tabular}{l|ccc}
    \toprule
   \multirow{2}{*}{\textsc{Retrieval method}}  & \multicolumn{3}{c}{\textsc{Accuracy}} \\
   &\textsc{Target}&\textsc{Bias}&\textsc{Both}\\
    \midrule
   \textsc{Bing} & \textbf{89.10} & 48.63 & \textbf{42.82} \\
     \textsc{CLIP + CC12M}& 77.66 & \textbf{50.29}& 38.05\\
     \textsc{CLIP + ImageNet} & 86.86 & 47.31 & 40.35\\
    \bottomrule
    \end{tabular}}}
\end{table}

In both~\cref{tab:eval4_fac,tab:eval4_ic}, we can see that Bing-retrieved images are generally more likely to contain the right target class, while CLIP-retrieved images are more likely to contain the right bias class. According to the VQA, on average, Bing is more accurate than CLIP-retrieval. However, as discussed in~\cref{sec:limitations}, there is still a lot of room for improvement, as the accuracy for ``both" (the case where the image contains the right target class and the right bias class) is below 50\% for all methods, according to the VQA.
For additional details about the VQA model, please refer to~\cref{sec:vqa_details}, where the accuracy of the VQA itself is measured.

\clearpage

\section{Details and comparisons of VQA models}
\label{sec:vqa_details}

\begin{table}[ht!]
\centering
\caption{VQA models chosen for comparison.}
\label{tab:vqa-specs}
\scalebox{0.7}{
  \begin{tabular}{@{}llccc@{}}
\toprule
\textsc{Model} &\textsc{Version} & \textsc{Params} & \textsc{Size (GB)} & \textsc{Released} \\ \midrule
\textsc{LLaVA-1.5} & \texttt{1.5-13b-hf}            & 13B     & 26.69     & 10/2023      \\
\textsc{LLaVA-NeXT}&\texttt{v1.6-vicuna-13bf-hf}    & 13B     & 26.69     & 01/2024      \\
 \bottomrule
\end{tabular}}
\end{table}

For the results presented in~\cref{sec:eval_setting_2}, we chose to pseudo-label the images with LLaVA-1.5-13B~\cite{liu2023llava, liu2023improvedllava}, which we found to be significantly faster than LLaVA-NeXT~\cite{liu2024llavanext}, with comparable accuracy for our use case (see~\cref{tab:vqa-specs,tab:vqa_eval}).

To label images with \ours-proposed biases, we use multiple-choice questions about bias attributes, with answer choices representing the proposed bias classes.
In the case of B2T, we ask binary yes-no questions about the presence of keywords associated to bias in images.

\begin{table}[ht!]
    \centering
    \caption{Comparing VQA models across various tasks and datasets. \textsc{Acc.} is the accuracy, \textsc{BM} the informedness metric, and \textsc{time} the run time per image (in ms).}
    \label{tab:vqa_eval}
    \scalebox{0.7}{
    \begin{tabular}{l| ccccc| lcc| ccccc}
    \toprule
     & \multicolumn{5}{c|}{\textsc{CelebA face attributes}} & \multicolumn{3}{c|}{\textsc{ImageNet classes}} & \multicolumn{5}{c}{\textsc{ImageNet-X factors}}\\
    \textsc{Model} & \textsc{Acc.} & \textsc{TPR} & \textsc{TNR} & \textsc{BM} & \textsc{Time} & & \textsc{Acc.} & \textsc{Time} & \textsc{Acc.} & \textsc{TPR} & \textsc{TNR} & \textsc{BM} & \textsc{Time}\\
    \midrule
        LLaVA-1.5 & 76.50 & \textbf{75.18} & 76.88 & \textbf{0.521} & \textbf{120} & & \textbf{85.29} & \textbf{105} & 39.09 & \textbf{85.15} & 30.28 & 0.154 & \textbf{101} \\
        LLaVA-NeXT & \textbf{78.80} & 66.57 & \textbf{82.29} & 0.489 & 217 & & 71.20 & 433 & \textbf{48.33} & 72.20 & \textbf{43.76} & \textbf{0.160} & 592 \\
    \bottomrule
    \end{tabular}}
\end{table}

In~\cref{tab:vqa_eval}, we compare the performance of LLaVA-1.5 and LLaVA-NeXT (both 13B versions) to predict binary face attributes on Celeba (left), target classes on ImageNet (center), and binary factors on ImageNet-X (right).

Because of class imbalance, we choose to also show true positive rate (TPR) and true negative rate (TNR) when classifying binary face attributes on CelebA or binary factors on ImageNet-X. These metrics are combined into (bookmaker) informedness (BM), also known as Youden's J statistic, defined as $BM = TPR + TNR - 1$. Informedness is proportional to balanced accuracy and is considered to be a more appropriate metric to assess the random guessing level of a classifier~\cite{chicco2021mcc}.

Overall, we found LLaVA-1.5 and LLaVA-NeXT to have comparable performance on the two datasets, but noted that LLaVA-1.5 had a tendency to reply more positively than LLaVA-NeXT (especially visible when looking at TPR and TNR). Moreover, we observed that, in our case, LLaVA-1.5 was better at following instructions (giving an answer within the given choices) than LLaVA-NeXT, with LLaVA-NeXT being more subject to hallucinations.

\clearpage

\section{Retrieved images diversity}
\label{sec:retrieval_diversity}

The quality of retrieved images plays a crucial role in \ours’s performance. While retrieval accuracy is important, the diversity of retrieved samples for each caption is equally critical. Low diversity can lead to overly homogeneous image sets that do not accurately capture the variability of the intended bias attribute, potentially skewing bias scores or reducing the robustness of bias detection.

To quantify retrieval diversity, we compute the average pairwise CLIP embedding similarity between all the retrieved images for a given target class (\textsc{Target Class Sim.}), as well as for each caption (\textsc{Bias Class Sim.}). A higher average similarity indicates less diversity (more homogeneous images), while a lower value indicates more varied visual content. We compare diversity scores for images retrieved from all retrieval sources, including domain-specific datasets, to assess which source provides more diverse image sets.

\begin{table}[ht!]
    \caption{Average cosine similarity between retrieved image embeddings across both tasks.}
    \label{tab:retrieval_diversity}
    \centering
    \scalebox{0.7}{
    \begin{tabular}{l|l c c c |c c c }
    \toprule
    & \multicolumn{4}{c|}{\textsc{Face attribute classification}} & \multicolumn{3}{c}{\textsc{Image classification}}\\
    \textsc{Retrieval source} & & \textsc{Bing} & \textsc{cc12m} & \textsc{CelebA} & \textsc{Bing} & \textsc{cc12m} & \textsc{ImageNet}\\
    \midrule
    \textsc{Target Class Sim.} 
    &
    & \textbf{0.60} 
    & 0.65 
    & 0.66
    & \textbf{0.75} 
    & \textbf{0.75} 
    & 0.78\\
    \textsc{Bias Class Sim.} 
    &
    & \textbf{0.65} 
    & 0.72 
    & 0.71
    & \textbf{0.77}
    & 0.78 
    & 0.80\\
    \bottomrule
    \end{tabular}
    }
\end{table}

As shown in~\cref{tab:retrieval_diversity}, Bing retrieval produces the most diverse image sets, followed by CC12M, followed by domain-specific datasets (CelebA and ImageNet). This confirms that web-scale retrieval sources provide richer visual variety. We hypothesize that this higher diversity contributes to better detection of complex or subtle biases and may explain why Bing-based retrieval often yields slightly better bias discovery performance in our experiments. These findings suggest that diversity-aware filtering or weighting strategies could be beneficial future improvements for bias assessment pipelines like \ours.

\clearpage

\section{Prompts used for LLM bias proposal}
\label{sec:prompts}
In the following, we describe the textual elements of \ours: the user input, the bias generation prompts, and the caption ones. 

\mypar{User input.} The only input needed from the user is a textual description of the task, including the target classes. We provide the descriptions that we used for our two example classification tasks in~\cref{fig:fac_description,fig:ic_description}. New task descriptions can easily be added to detect biases for different classification tasks.

\begin{figure}[ht!]
    \begin{lstlisting}
    In the domain of face attribute classification, models are trained to identify
    attributes from images of human faces. These attributes can include, but are not
    limited to, age, gender, expression, presence of accessories (e.g., glasses, hats),
    facial hair, skin tone, and any distinctive features.
    \end{lstlisting}
    \caption{Task description for face attribute classification.}
    \label{fig:fac_description}
\hfill
\vfill
    \begin{lstlisting}
    In the domain of image classification, models are trained to assign a label or category
    to an entire image, identifying objects, scenes, or concepts based on visual content.
    \end{lstlisting}
    \caption{Task description for image classification.}
    \label{fig:ic_description}%
\end{figure}

\mypar{Bias generation prompts.}
As explained in~\cref{sec:method-biases}, different biases are generated for each target class. The LLM is prompted in a ``chat completion" mode using the ChatML format~\cite{ChatML}.
First, a \textit{system} prompt (\cref{fig:bias_system_prompt}) is given with general instructions about the task at hand. Then, a \textit{user} prompt is given, containing information about the task (name and description), the target attribute, the target class, as well as some additional instructions (\cref{fig:bias_user_instructions}) for additional guidance. The expected response format is also provided to the LLM (\textit{JSON} Schema).

\begin{figure}[ht!]
    \begin{lstlisting}
    You are a helpful intelligent assistant knowledgeable about computer vision. Our
    objective is to identify potential visual biases in pre-trained computer vision
    classifier models. Given the description of a computer vision task, the name of a
    target attribute, and the name of the class that we are trying to identify, generate
    a list of potential visual biases that a classifier may have. These biases must be
    identifiable in images and should reflect common issues that can arise from the
    training data or model architecture. 

    Instructions:
    1. Generate a list of visually-identifiable bias attributes that could influence the
    performance of a pre-trained classifier for a given task.
    2. For each bias attribute, provide a list of bias classes that represent all the
    potential values of this attribute.
    3. Output the list of bias attributes and their classes in JSON format.
    \end{lstlisting}
    \caption{System prompt for bias generation.}
    \label{fig:bias_system_prompt}

\hfill
\vfill

    \begin{lstlisting}
    Think about what characteristic or feature of the image could impact the performance
    of the model. Think about potential spurious correlations and potential failure types.
    \end{lstlisting}
    \caption{Additional instructions given in the user prompt for bias generation.}
    \label{fig:bias_user_instructions}
\end{figure}

\mypar{Caption template prompts.}
As explained in~\cref{sec:method-retrieval}, the first step is to generate a caption template for the task, which will then be adapted by the LLM to fit different combinations of target and bias classes. As is done for bias generation, a system prompt with general instructions is first given (\cref{fig:template_system_prompt}), and a user prompt is given in a second phase, containing the task name, the task description, and additional instructions (\cref{fig:template_user_instructions}).

\begin{figure}[ht!]
    \begin{lstlisting}
    You are a large language model designed to generate simple captions for images related
    to a specific computer vision task. Given the description of a computer vision task,
    your task is to generate a simple caption template that would work for any image
    relevant to this task. This template will be used at the beginning of every caption, so
    keep it general. Think about general characteristics that apply to all images of this
    task. The template should only correspond to the beginning of a sentence, and end with
    "{}", a placeholder that will be personalized in the future to generate all captions for
    this task.
    \end{lstlisting}
    \caption{System prompt for caption template generation.}
    \label{fig:template_system_prompt}
\hfill
\vfill
    \begin{lstlisting}
    Think about the type of image (e.g., a photo) and what it should contain. Keep the
    template simple and general. Do not explicitly mention the task in the template. Avoid
    unnecessary words and do not make sentences.
    \end{lstlisting}
    \caption{Additional instructions given in the user prompt for caption template generation.}
    \label{fig:template_user_instructions}
\end{figure}

\mypar{Caption generation prompts.}
Finally, captions are generated for each combination of target and bias classes. In practice, the LLM is only prompted for each bias attribute (for each target class), and produces the captions for all the corresponding bias classes at once. This reduces the number of calls to the LLM, allows the captions to be more consistent across bias classes, and helps the generation process by providing more context to the LLM. Again, a system prompt (\cref{fig:caption_system_prompt}) is initially given to the LLM, before a user prompt containing information about the task (name and description), the target class, the bias attribute and bias classes, the caption template, as well as additional instructions (\cref{fig:caption_user_instructions}) is also given.

\begin{figure}[ht!]
    \begin{lstlisting}
    You are a large language model designed to generate simple captions for images related
    to a specific computer vision task. Your objective is to generate captions as simple
    combinations of attributes to retrieve images that match the task as well as the
    attributes. Given the description of a computer vision task, a target class, a bias
    attribute, and bias classes for this attribute, your task is to generate simple captions
    that describe images that combine the given elements. To help generate relevant captions,
    a template is also provided, you can adapt the template in any way you want, even change
    it if necessary. The captions should make grammatical and logical sense, be relevant to
    the task, and always combine the given target class and bias class. Please avoid using
    negations, provide the opposite meaning of the negated word.
    \end{lstlisting}
    \caption{System prompt for caption generation.}
    \label{fig:caption_system_prompt}
\hfill
\vfill
    \begin{lstlisting}
    Do not introduce any new bias in the captions. Do not add new attributes. The only
    attributes included in the caption should be the given target class and the given
    bias class.
    \end{lstlisting}
    \caption{Additional instructions given in the user prompt for caption generation.}
    \label{fig:caption_user_instructions}
\end{figure}

\clearpage

\section{Comparison between different LLMs for bias proposal}
\label{sec:llm-comparison}
To generate potential biases, we have chosen to compare several lightweight quantized versions of recent state-of-the-art LLMs of comparable sizes, \ie, Gemma~\cite{team2024gemma}, Llama~\cite{dubey2024llama}, and Phi~\cite{abdin2024phi}. General specifications are shown in~\cref{tab:llm-specs}.

\begin{table}[ht!]
\centering
\caption{Lightweight LLMs chosen for comparison.}
\label{tab:llm-specs}
\scalebox{0.7}{
  \begin{tabular}{@{}llcccc@{}}
\toprule
\textsc{Model} &\textsc{Version}                          & \textsc{Params} & \textsc{Quantization} & \textsc{Size (GB)} & \textsc{Released} \\ \midrule
\textsc{Gemma} & \texttt{2-9b-it}              & 9B     & \texttt{Q6\_K\_L}     & 7.81      & 06/2024      \\
\textsc{Llama}&\texttt{3.1-8B-Instruct}    & 8B     & \texttt{Q8\_0}        & 8.54      & 07/2024      \\
\textsc{Phi}&\texttt{3-medium-128k-instruct} & 14B    & \texttt{Q4\_K\_M}     & 8.57      & 05/2024      \\ \bottomrule
\end{tabular}}
\end{table}

\begin{table}[ht!]
\centering
\caption{Comparison between lightweight LLMs for bias proposal. $\lvert \mathcal{B} \rvert$ is the average number of bias attributes per class, $\lvert \mathcal{B} \rvert$ the average number of bias classes per class, and \textsc{time} is the average run time in seconds. }
\label{tab:llm-comparison}
\scalebox{0.7}{
\begin{tabular}{llccccc}
\toprule
 \textsc{Task} & \textsc{Model} &$\lvert \mathcal{B} \rvert$ & $\lvert \mathbf{B} \rvert$  & ${\lvert \mathbf{B} \rvert}/{\lvert \mathcal{B} \rvert}$ & \textsc{time} & \textsc{time}$/{\lvert \mathbf{B} \rvert}$\\ %
\midrule
\multirow{3}{*}{\textsc{CelebA}} %
& \textsc{Gemma} & 7.58 & 25.68 & \textbf{3.39} & 13.76 & 0.54\\
 & \textsc{Llama} & \textbf{11.00} & \textbf{35.45} & 3.26 & 11.60 & 0.33\\
 & \textsc{Phi} & 6.28 & 20.90 & 3.28 & \textbf{4.11} & \textbf{0.20}\\
\hline
\multirow{3}{*}{\textsc{ImageNet}} & \textsc{Gemma} & 6.16 & 23.17 & \textbf{3.76} & 12.74 & 0.55\\
 & \textsc{Llama} & \textbf{9.74} & \textbf{28.72} & 2.95 & 8.72 & 0.30\\
 & \textsc{Phi} & 6.34 & 18.62 & 2.94 & \textbf{3.73} & \textbf{0.20}\\
\bottomrule
\end{tabular}
}
\end{table}

In Table \ref{tab:llm-comparison}, we show the average number of proposed bias attributes ($\lvert \mathcal{B} \rvert$) and corresponding bias classes per target class ($\lvert \mathbf{B} \rvert$), as well as the execution time per target class (\textsc{time}), for both face attribute classification and image classification.
With our prompting strategy, Llama was the LLM giving us the largest number of potential biases, while Phi was surprisingly fast. On average, Gemma proposes more bias classes per bias attribute, but is slower than both Llama and Phi.

Qualitatively, proposed biases are fairly similar between the three tested LLMs. We provide some examples of bias that were proposed by Gemma, LLama, and Phi in~\cref{fig:gemma_bias,fig:llama_bias,fig:phi_bias}, respectively, for the \textit{smiling} target attribute of CelebA. On average, we have found the biases proposed by Llama to be slightly more relevant than the ones proposed by Gemma and Phi, which drove our decision to choose it for \ours.

\begin{figure}[ht!]
    \begin{lstlisting}[language=json]
 [{"bias attribute": "Lighting",
   "bias classes": ["Bright", "Dim", "Shadowed"]},
  {"bias attribute": "Pose",
   "bias classes": ["Front-facing", "Profile", "Three-quarter"]},
  {"bias attribute": "Facial Expression Context",
   "bias classes": ["Happy", "Sad", "Neutral", "Angry", "Surprised"]},
  {"bias attribute": "Image Quality",
   "bias classes": ["High Resolution", "Low Resolution", "Blurry", "Distorted"]},
  {"bias attribute": "Cultural Background",
   "bias classes": ["Western", "Eastern", "African", "Other"]},
  {"bias attribute": "Age",
   "bias classes": ["Young", "Adult", "Elderly"]}]
    \end{lstlisting}
    \caption{Example of biases proposed by Gemma, for the \textit{smiling} target attribute of CelebA.}
    \label{fig:gemma_bias}
\end{figure}

\begin{figure}[ht!]
    \begin{lstlisting}[language=json]
 [{"bias attribute": "Lighting",
   "bias classes": ["Bright", "Dim", "Shaded"]},
  {"bias attribute": "Facial Expression",
   "bias classes": ["Smiling", "Neutral", "Frowning"]},
  {"bias attribute": "Glasses",
   "bias classes": ["Present", "Absent"]},
  {"bias attribute": "Hats and Headwear",
   "bias classes": ["Present", "Absent"]},
  {"bias attribute": "Facial Hair",
   "bias classes": ["Present", "Absent"]},
  {"bias attribute": "Skin Tone",
   "bias classes": ["Fair", "Medium", "Dark"]},
  {"bias attribute": "Age",
   "bias classes": ["Young", "Old"]},
  {"bias attribute": "Image Quality",
   "bias classes": ["High Resolution", "Low Resolution"]},
  {"bias attribute": "Camera Angle",
   "bias classes": ["Frontal", "Profile"]},
  {"bias attribute": "Background Clutter",
   "bias classes": ["Clean", "Cluttered"]}]
    \end{lstlisting}
    \caption{Example of biases proposed by Llama, for the \textit{smiling} target attribute of CelebA.}
    \label{fig:llama_bias}
\end{figure}

\begin{figure}[ht!]
    \begin{lstlisting}[language=json]
 [{"bias attribute": "Facial Expression",
   "bias classes": ["Smiling", "Neutral", "Frowning"]},
  {"bias attribute": "Lighting Conditions",
   "bias classes": ["Bright Light", "Dim Light", "Backlight", "Shadowed Face"]},
  {"bias attribute": "Facial Accessories",
   "bias classes": ["Glasses", "Hats", "Masks", "None"]},
  {"bias attribute": "Skin Tone",
   "bias classes": ["Light Skin", "Dark Skin", "Tanned Skin", "Pale Skin"]},
  {"bias attribute": "Facial Hair",
   "bias classes": ["Beard", "Mustache", "Clean-Shaven", "None"]},
  {"bias attribute": "Age Group",
   "bias classes": ["Young", "Middle Age", "Older Adults"]}]
    \end{lstlisting}
    \caption{Example of biases proposed by Phi, for the \textit{smiling} target attribute of CelebA.}
    \label{fig:phi_bias}
\end{figure}

\clearpage

\section{Embedding-based bias matching details and examples}
\label{sec:embedding-matching}
The proposed biases may be names that do not match those in the ground truth (\eg, even due to synonyms, such as ``male" and ``man"). Thus, for the evaluation based on ground-truth annotations presented in~\cref{sec:eval_setting_1}, %
we match proposed and ground-truth biases %
using cosine similarity of their respective SBERT embeddings.

In practice, we found similarity scores based on embeddings of single words to be unreliable. The embeddings were actually computed on captions containing the bias attributes/classes, \ie, ``A photo of a young person", instead of ``Young".

\begin{figure}[ht!]
  \centering
  \includegraphics[width=\textwidth]{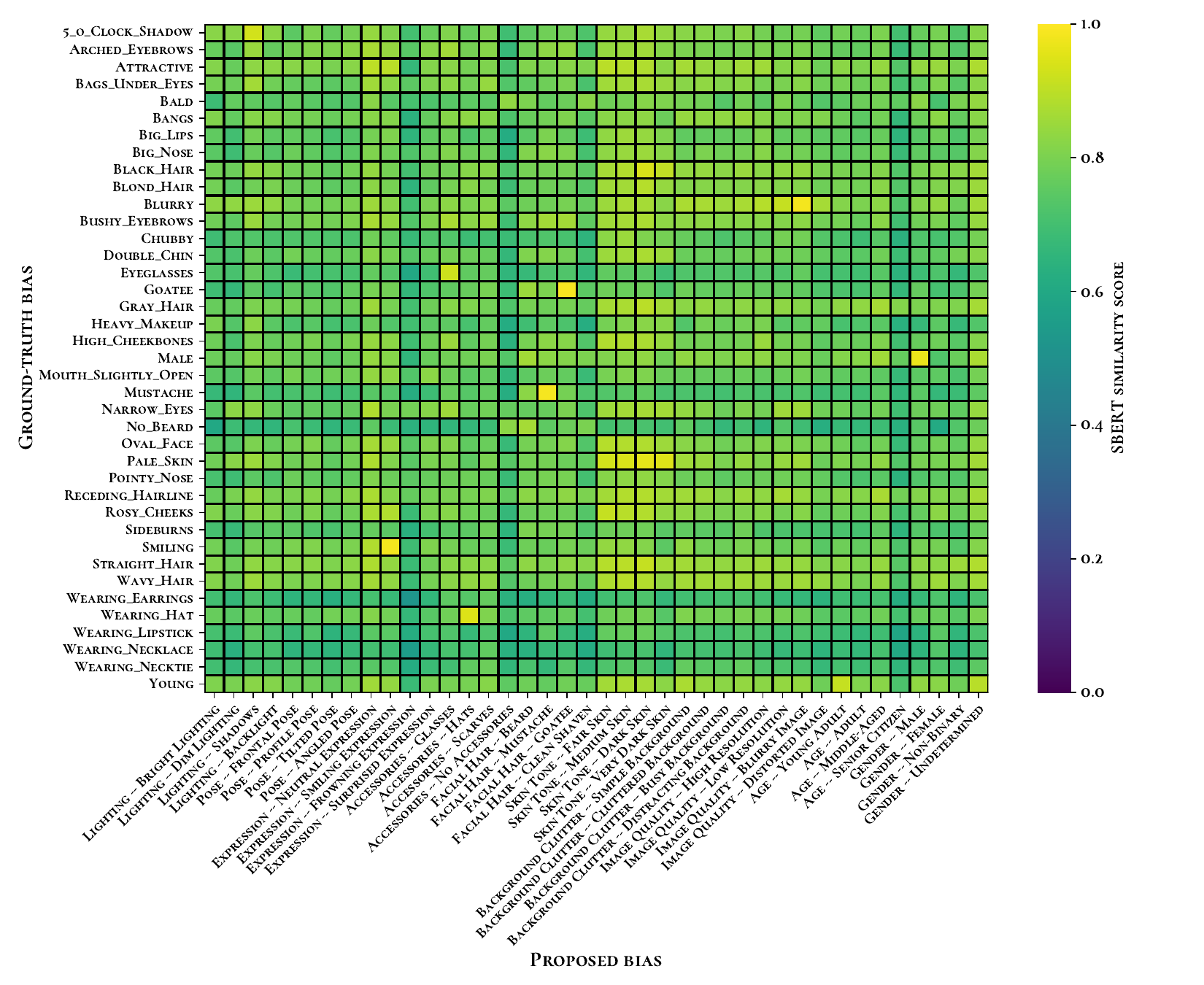}
  \caption{Example of SBERT similarity scores between ground-truth attributes (used as ground-truth biases) and biases proposed by \ours for the \textit{brown hair} attribute on CelebA.}
  \label{fig:matching_matrix}
\end{figure}

In~\cref{fig:matching_matrix}, we show an example of similarity scores between ground-truth attributes (used as ground-truth biases) and bias proposed by \ours for the \textit{brown hair} attribute on CelebA.
First, a similarity threshold is defined as a minimum score for two biases to be matched. This threshold was set the 0.9 for the results we present in~\cref{sec:eval_setting_1}. Matching biases is then an iterative process, where the most similar pair is matched and removed from the similarity matrix, until no pairs above the similarity threshold are left.

\begin{table}%
\centering
\caption{Examples of similarity scores (\textsc{Sim. Score}) between ground-truth attributes (used as ground-truth biases) and bias proposed by \ours for the \textit{attractive} attribute on CelebA. %
}
\label{tab:matching_scores}
\scalebox{0.7}{\begin{tabular}{@{}lllc@{}}
\toprule
\multirow{2}{*}{\textsc{Ground-truth}}   & \multicolumn{2}{c}{\textsc{Proposed}}  &  \textsc{Sim.}  \\
  & \textsc{Attribute}  & \textsc{Class} &  \textsc{Score}\\ \midrule
Smiling             & Facial Expression       & Smiling             & 0.98\\
Heavy\_Makeup       & Makeup                  & Heavy               & 0.98\\
Young               & Age                     & Young               & 0.98\\
Eyeglasses          & Glasses                 & Present             & 0.93\\
Pale\_Skin          & Skin Tone               & Dark                & 0.88\\
Blurry              & Lighting                & Dim                 & 0.84\\
Rosy\_Cheeks        & Lighting                & Bright              & 0.84\\
Goatee              & Facial Hair             & Present             & 0.83\\
5\_o\_Clock\_Shadow & Lighting                & Shaded              & 0.81\\
\bottomrule
\end{tabular}}
\end{table}

In ~\cref{tab:matching_scores}, we show examples of similarity scores between ground-truth attributes (used as ground-truth biases) and bias proposed by \ours for the \textit{attractive} attribute on CelebA. We can see that a similarity score of 0.95 would have missed a true match, while a similarity score of 0.8 would have been too low and would have resulted in too many false matches.

We show additional quantitative results varying the similarity threshold in~\cref{sec:additional_quantitative} (\cref{tab:eval_celeba_gt_s08,tab:eval_celeba_gt_s095,tab:eval_inx_gt_s08,tab:eval_inx_gt_s095}).

\clearpage

\section{Ground-truth bias matrices visualizations}
\label{sec:gt-biases-visualization}
We follow the definition of bias given in~\cref{sec:def} to compute ground-truth biases, which are then used for the evaluation based on ground-truth annotations presented in~\cref{sec:eval_setting_1}.

In this section, we show the ground-truth bias matrices for FaceXFormer on CelebA, as well as {ResNet50\_V2}, {ResNet101\_V2},  {ResNet152\_V2}, and {ViT\_B\_16\_SWAG\_E2E\_V1} on ImageNet-X.

\begin{figure}[ht!]
  \centering
  \includegraphics[width=\textwidth]{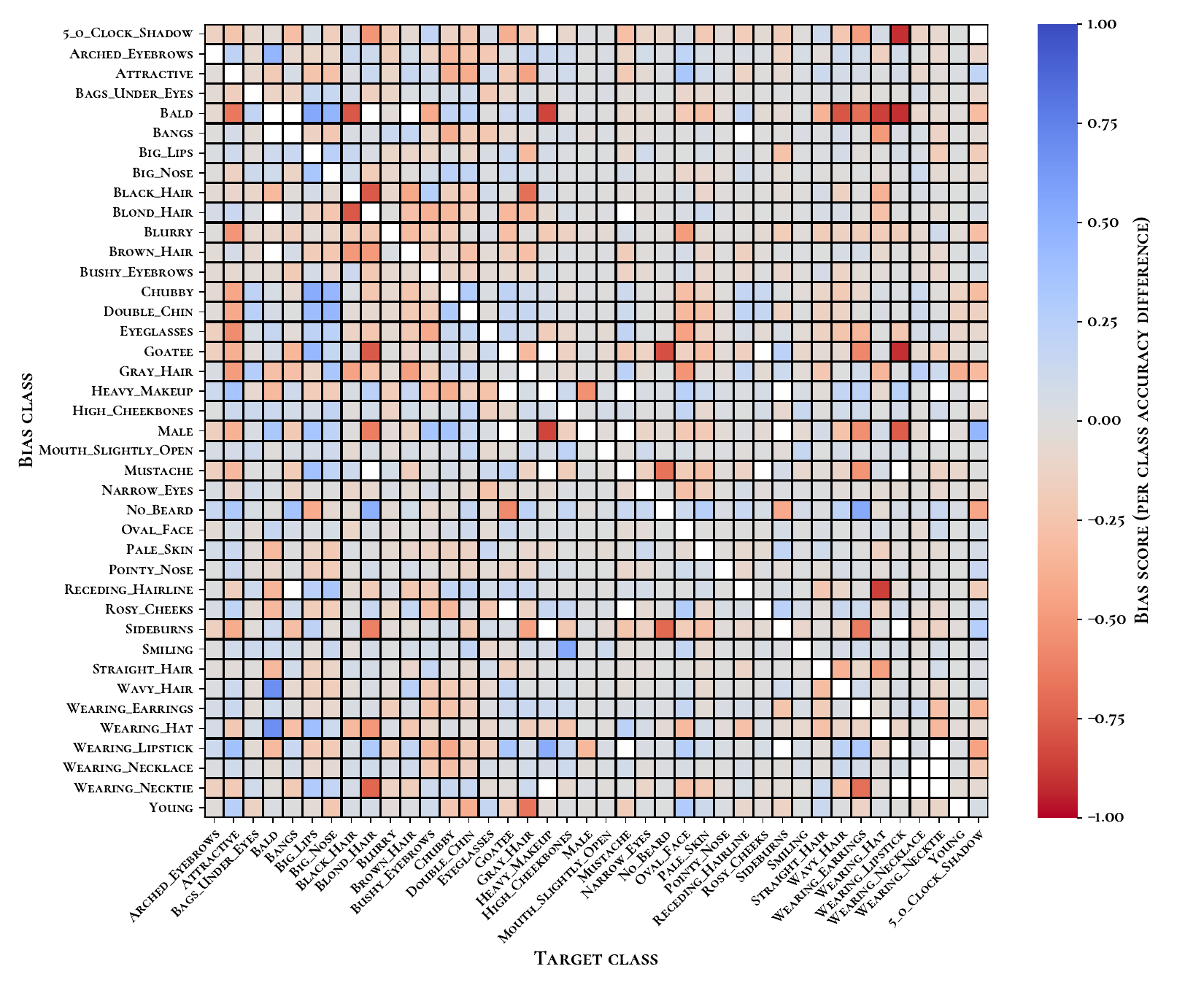}
  \caption{Ground-truth bias matrix of FaceXFormer on CelebA.}
  \label{fig:celeba_bias_matrix}
\end{figure}

\begin{figure}[ht!]
  \centering
  \includegraphics[width=\textwidth]{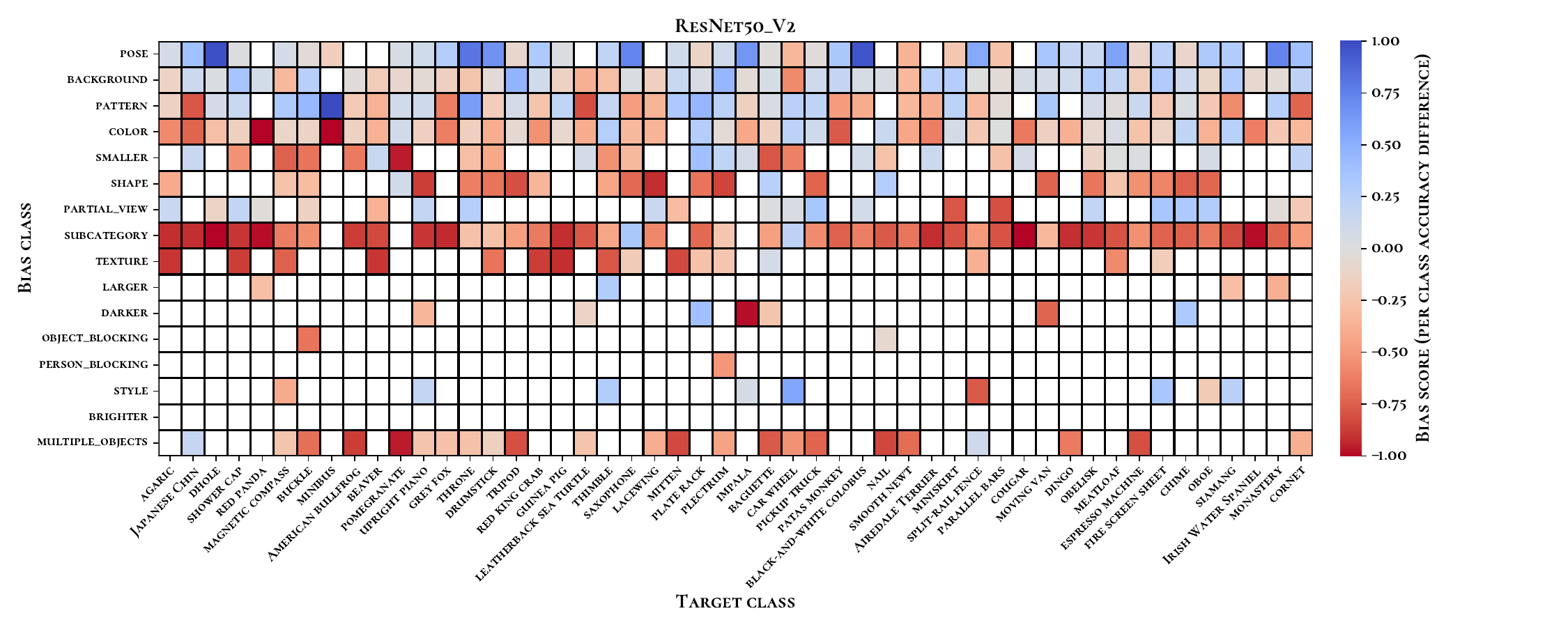}
  \caption{Ground-truth bias matrix of {ResNet50\_V2} for the 50 classes with the strongest biases on ImageNet-X.}
  \label{fig:inx_top50_bias_matrix_rn50}
\end{figure}

\begin{figure}[ht!]
  \centering
  \includegraphics[width=\textwidth]{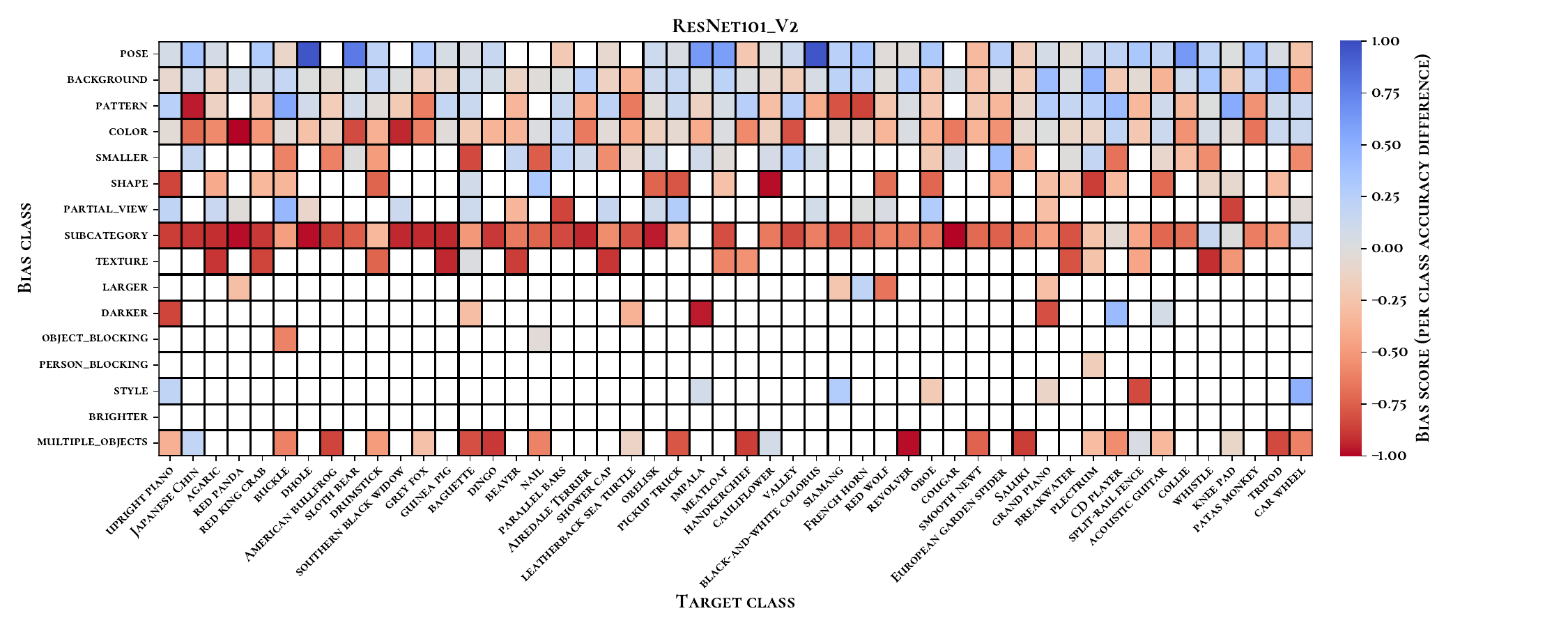}
  \caption{Ground-truth bias matrix of {ResNet101\_V2} for the 50 classes with the strongest biases on ImageNet-X.}
  \label{fig:inx_top50_bias_matrix_rn101}
\end{figure}

\begin{figure}[ht!]
  \centering
  \includegraphics[width=\textwidth]{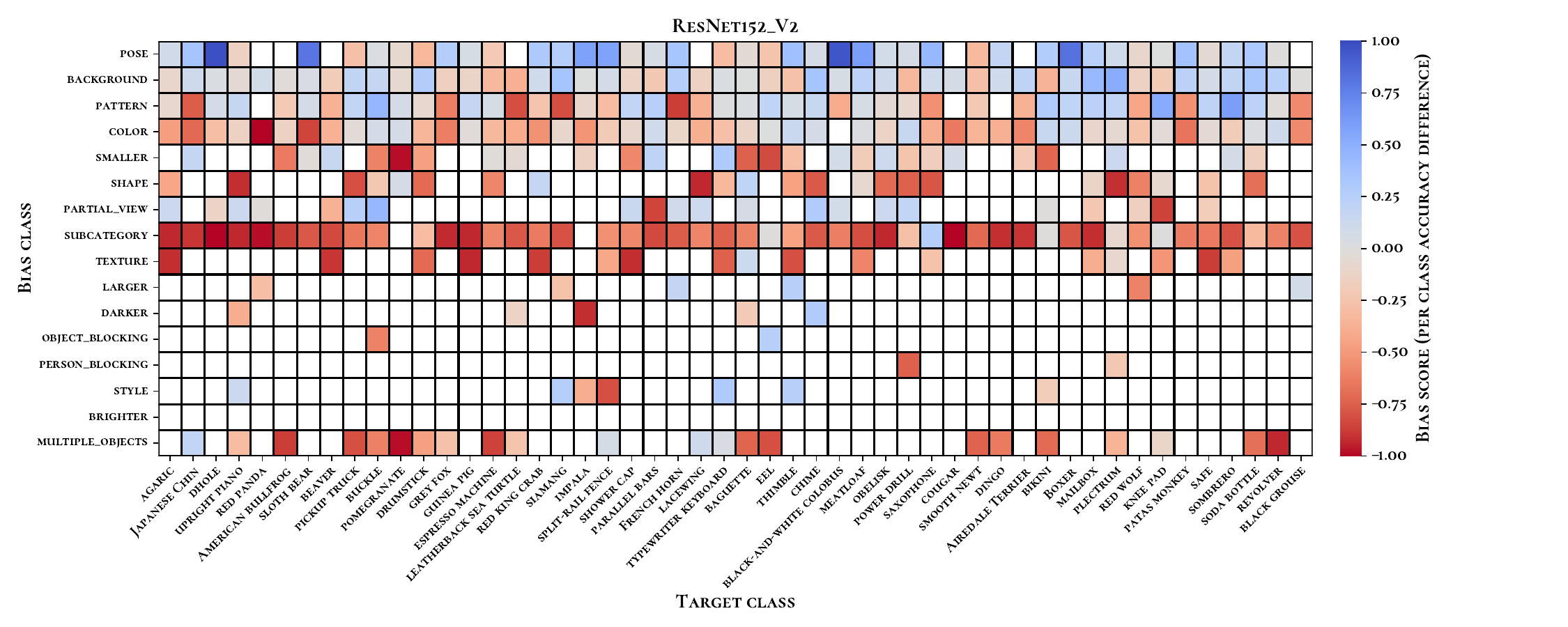}
  \caption{Ground-truth bias matrix of {ResNet152\_V2} for the 50 classes with the strongest biases on ImageNet-X.}
  \label{fig:inx_top50_bias_matrix_rn152}
\end{figure}

\begin{figure}[ht!]
  \centering
  \includegraphics[width=\textwidth]{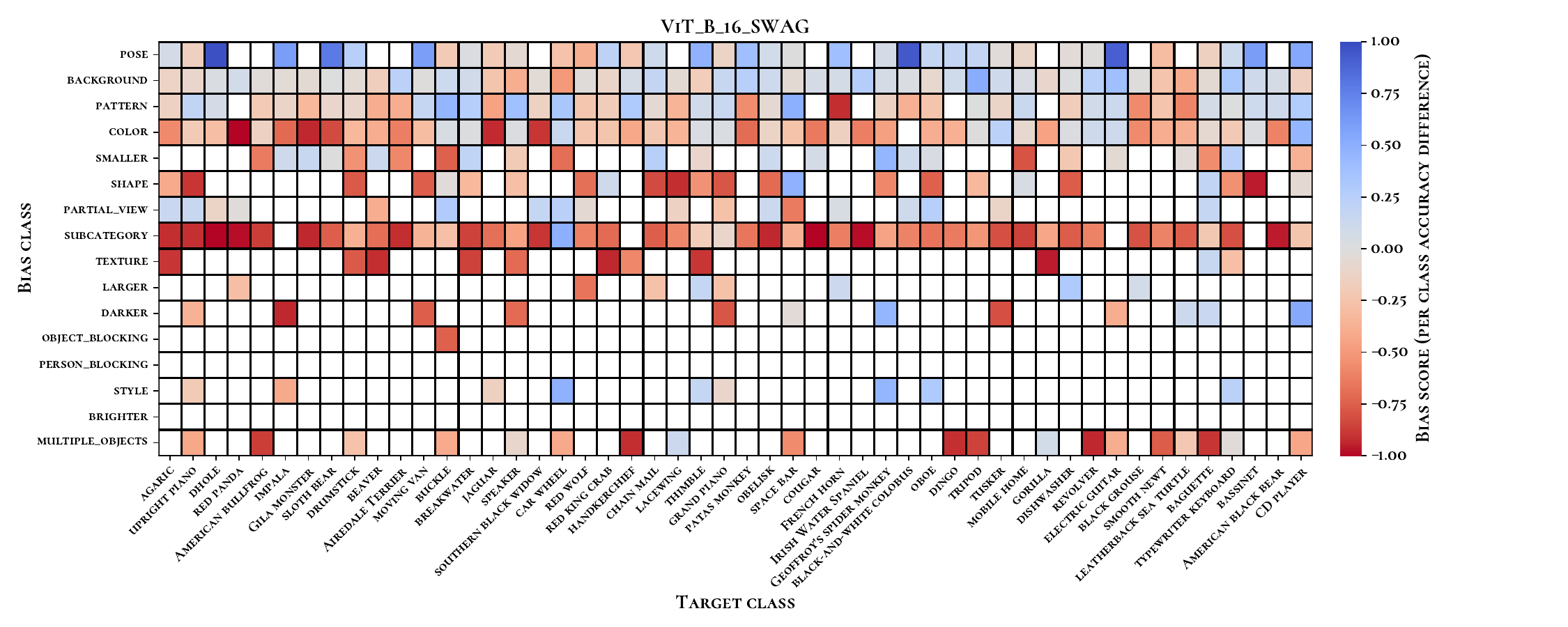}
  \caption{Ground-truth bias matrix of {ViT\_B\_16\_SWAG\_E2E\_V1} for the 50 classes with the strongest biases on ImageNet-X.}
  \label{fig:inx_top50_bias_matrix_vit}
\end{figure}

In~\cref{fig:celeba_bias_matrix}, we show the ground-truth bias matrix of FaceXFormer for all classes on CelebA, where per the class accuracy is equivalent to the true positive rate for each binary attribute.

For ImageNet-X, as it would be impossible to see individual biases in full ground-truth bias matrices over all 1000 classes without zooming in, we provide bias matrices for the top 50 classes with the strongest biases for each model in~\cref{fig:inx_top50_bias_matrix_rn50,fig:inx_top50_bias_matrix_rn101,fig:inx_top50_bias_matrix_rn152,fig:inx_top50_bias_matrix_vit}.

For all ground-truth bias matrices, a positive bias score (blue color) indicates a higher accuracy when the bias is present, and a negative bias score (red color) indicates a lower accuracy when bias is present.
A white square indicates that the bias could not be measured because there was no example in the dataset.

\end{document}